\documentclass[journal]{IEEEtran}
\usepackage{graphicx}
\usepackage{subcaption}
\usepackage{multirow}
\usepackage{booktabs}
\usepackage[table]{xcolor}
\usepackage{color}
\usepackage{colortbl}
\usepackage{tabularx}
\usepackage{bm}
\usepackage{url}
\usepackage{stackengine}
\usepackage{float}
\usepackage{verbatim}
\usepackage{array}
\usepackage{cite}
\usepackage{algorithm}
\usepackage{algorithmic}
\usepackage[colorlinks,linkcolor=blue]{hyperref}
\usepackage{amsmath,amssymb} 
\usepackage{comment}
\usepackage{svg}
\usepackage{orcidlink}

\begin{document}
\title{A Survey of Sample-Efficient Deep Learning for Change Detection in Remote Sensing: Tasks, Strategies, and Challenges}

\author{
Lei Ding, 
Danfeng Hong, \IEEEmembership{Senior Member,~IEEE,}
Maofan Zhao, 
Hongruixuan Chen,
Chenyu Li, \IEEEmembership{Student Member,~IEEE,}
Jie Deng,
Naoto Yokoya, \IEEEmembership{Member,~IEEE,}
Lorenzo Bruzzone, \IEEEmembership{Fellow,~IEEE,}
and Jocelyn Chanussot,~\IEEEmembership{Fellow,~IEEE}

\thanks{This work was supported by the National Natural Science Foundation of China under Grant 42201443, Grant 42271350, and also supported by the International Partnership Program of the Chinese Academy of Sciences under Grant No.313GJHZ2023066FN.(\emph{Corresponding author: Danfeng Hong})}

\thanks{L. Ding is with the Aerospace Information Research Institute, Chinese Academy of Sciences, Beijing, and also with the Information Engineering University, Zhengzhou, China. (E-mail: dinglei14@outlook.com).}

\thanks{D. Hong is with the Aerospace Information Research Institute, Chinese Academy of Sciences, Beijing, 100094, China, and also with the School of Electronic, Electrical and Communication Engineering, University of Chinese Academy of Sciences, 100049 Beijing, China. (e-mail: hongdf@aircas.ac.cn)}

\thanks{M. Zhao and J. Deng are with the Aerospace Information Research Institute, Chinese Academy of Sciences, 100094 Beijing, China. (e-mail: mfzhao1998@163.com, dengjie@aircas.ac.cn)}

\thanks{C. Li is with the School of Mathematics and Statistics, Southeast University, 211189 Nanjing, China. (e-mail: lichenyu@seu.edu.cn)}

\thanks{H. Chen is with the Graduate School of Frontier Sciences, The University of Tokyo, Chiba 277-8561, Japan (e-mail: Qschrx@gmail.com).}

\thanks{N. Yokoya is with the Department of Complexity Science and Engineering, Graduate School of Frontier Sciences, The University of Tokyo, Chiba 277-8561, Japan, and also with the RIKEN Center for Advanced Intelligence Project, Tokyo 103-0027, Japan (e-mail: yokoya@k.u-tokyo.ac.jp).}

\thanks{L. Bruzzone is with the Department of Information Engineering and Computer Science, University of Trento, 38123 Trento, Italy (lorenzo.bruzzone@unitn.it).}

\thanks{J. Chanussot is with Univ. Grenoble Alpes, Inria, CNRS, Grenoble INP, LJK, Grenoble, 38000, France. (e-mail: jocelyn.chanussot@inria.fr).}

}

\markboth{GRSM}%
{Shell \MakeLowercase{\textit{et al.}}: Bare Demo of IEEEtran.cls for IEEE Journals}

\maketitle

\begin{abstract}
    In the last decade, the rapid development of deep learning (DL) has made it possible to perform automatic, accurate, and robust Change Detection (CD) on large volumes of Remote Sensing Images (RSIs). However, despite advances in CD methods, their practical application in real-world contexts remains limited due to the diverse input data and the applicational context. For example, the collected RSIs can be time-series observations, and more informative results are required to indicate the time of change or the specific change category. Moreover, training a Deep Neural Network (DNN) requires a massive amount of training samples, whereas in many cases these samples are difficult to collect. To address these challenges, various specific CD methods have been developed considering different application scenarios and training resources. Additionally, recent advancements in image generation, self-supervision, and visual foundation models (VFMs) have opened up new approaches to address the 'data-hungry' issue of DL-based CD. The development of these methods in broader application scenarios requires further investigation and discussion. Therefore, this article summarizes the literature methods for different CD tasks and the available strategies and techniques to train and deploy DL-based CD methods in sample-limited scenarios. We expect that this survey can provide new insights and inspiration for researchers in this field to develop more effective CD methods that can be applied in a wider range of contexts.
\end{abstract}
\emph{Index Terms}---remote sensing, change detection, deep learning, supervised learning, visual foundation model.


\section{Introduction}\label{sec_intro}

Over the last 10 years, the emergence and success of Deep Learning (DL) techniques \cite{hong2024spectralgpt} have significantly advanced the field of Change Detection (CD) in Remote Sensing Images (RSIs). DL-based CD enables data-driven learning of specific changes of interest and, as a result, facilitates accurate and fully automatic processing of vast amounts of data. State-Of-The-Art (SOTA) methods~\cite{chen2021remote, Liky2023Changer, ding2024samcd} have reached an accuracy exceeding 90\% in the $F_1$ metric across multiple benchmark datasets for CD, highlighting the remarkable identification capability of DL-based CD approaches.

Despite these advances, the translation of CD methods into practical real-world applications remains a significant challenge. This arises from the inherent diversity present in the input RSIs, as well as the wide variety of scenarios to conduct CD algorithms. For instance, the multi-temporal RSIs that CD methods process can exhibit significant heterogeneity or spatial misalignment \cite{shi2020change}, and more fine-grained information is required to indicate the time of change or the specific change category. This necessitates the development of CD methodologies that can operate effectively within such varied and intricate environments.

Moreover, training a robust Deep Neural Network (DNN) for CD requires extensive and accurately labeled datasets. In many real-world scenarios, the presence of such data is scarce. The construction of a CD training set requires the collection of RSIs with expansive region coverage and adequate temporal intervals to capture changes of interest \cite{shen2021s2looking}. For some small or rare types of change, it is often difficult to collect a sufficient number of training samples. This poses negative impacts on the efficacy and generalization of DL-based CD approaches.

\begin{figure}[!t]
	\begin{center}
    \includegraphics[width = 0.5\textwidth]{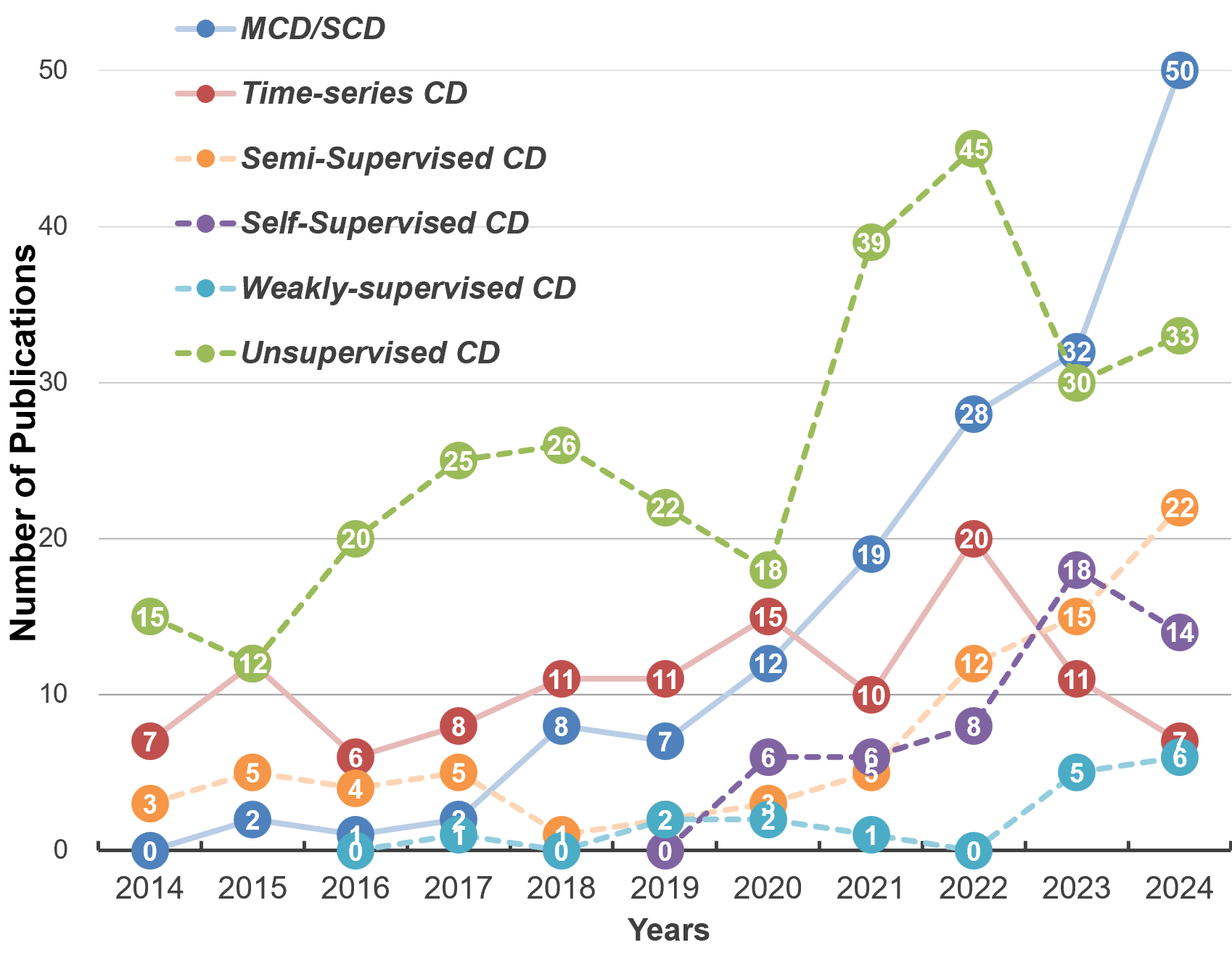}
	\end{center}
	\caption{The number of literature publications associated with different CD topics over the past 10 years. Solid lines present different CD tasks, while the dashed lines indicate different supervision strategies.}
	\label{fig.paper_num}
\end{figure}

In response to these challenges, researchers have developed a variety of specialized CD methodologies that are customized to specific application contexts and training limitations. These methodologies encompass various subdivided CD tasks, each designed to meet the unique demands of a particular scenario. Concurrently, innovative training techniques and strategies have been introduced to mitigate the issue of 'data-hungry' in training DNNs for CD. By exploring the underlying semantic context and multi-temporal correlations that are inherent to RSIs, the demand for extensive training labels can be reduced. Based on the different levels of supervision signals introduced, DL-based CD methods can be divided into several categories, such as fully supervised, semi-supervised, self-supervised, weakly supervised, and unsupervised. To display the dynamics in recent CD-related studies, in Fig.\ref{fig.paper_num} we present the number of publications associated with different CD tasks and supervision strategies. The statistics are obtained through a search at \textit{Web of Science} \footnote{https://www.webofscience.com/wos/} using related keywords while filtering the metadata to exclude those irrelevant to \textit{remote sensing}. One can observe that there has been a rapid growth of interest in several CD topics, including multi-class CD, self-supervised CD, and semi-supervised CD. Additionally, some incomplete supervision settings have been rarely studied until very recent years (e.g., weakly supervised CD). These statistics indicate a trend of research focus in recent studies: as fully-supervised CD has already reached a high level of accuracy, an increasing number of investigations are being conducted on more challenging CD topics with incomplete supervision setups \cite{cheng2023change}.

In light of these developments, there is a pressing need to comprehensively review and analyze the recent research on DL-based CD methods, particularly those tailored to diverse applicational contexts and incomplete supervision circumstances. This review aims to fill this gap by providing a detailed examination of the literature on CD tasks, which have been partitioned into specialized domains to address the unique challenges of each setting. In doing so, we expect to provide an in-depth understanding of the techniques and strategies employed to train and deploy DNN-based CD methods in real-world scenarios. Furthermore, we seek to identify gaps in the existing literature and highlight areas for future research, thus contributing to the multifaceted advancement and broader application of CD methodologies.

\section{CD Tasks}\label{sec_outline}

\begin{figure*}[t]
\centering
    \includegraphics[width=1\linewidth]{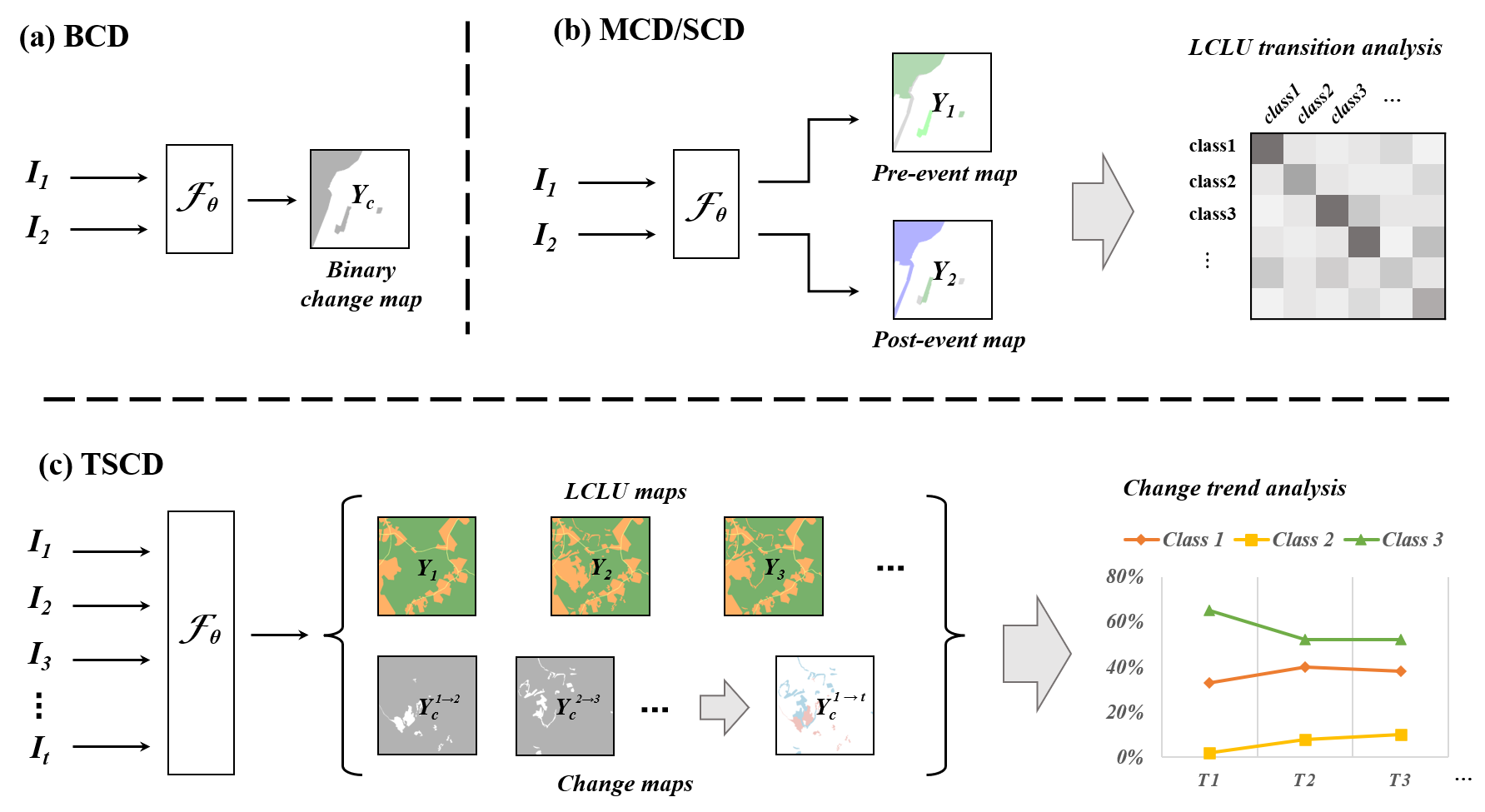}
    \caption{A comparison between (a) BCD, (b) MCD/SCD, and (c) TSCD. The color regions in $Y_1, Y_2, Y_3$ and $Y_c^{1 \rightarrow t}$ indicate the pre-defined LCLU/change categories.}
    \label{Fig.tasks}
\end{figure*}

According to the granularity of results and the type of input images, CD in RSIs can be further divided into various sub-categories, including Binary CD (BCD), Multi-class CD/Semantic CD (MCD/SCD), and Time-series CD (TSCD). Fig.\ref{Fig.tasks} presents an overview of these different tasks. In the following, we summarize the benchmarks, applicational scope, and representative works related to each CD task.

\subsection{Binary CD}

\textbf{Background:} BCD has been the most extensively studied CD task in the past few decades. Unless otherwise specified, BCD is also commonly abbreviated as CD in literature. As BCD has been comprehensively reviewed in existing literature, here we only provide a brief summary of the typical paradigms and representative work.

In the initial stages, DL-based BCD was seen as a segmentation task, where UNet-like Convolutional Neural Networks (CNNs) are employed to directly segment changes \cite{peng2019end}. Let $I_1$ and $I_2$ denote a pair of RSIs obtained on the dates $t_1$ and $t_2$, respectively. The general function of CD can be represented as:

\begin{equation}
    \mathcal{F}_{\theta}(I_1, I_2) = Y_c,
\end{equation}
where $Y_c$ is the predicted change map, $\mathcal{F}$ is the mapping function of a DNN with the trained parameters $\theta$. Differently, Daudt et al. \cite{daudt2018fully} proposed to first extract the temporal features, then model the change representations:
\begin{equation} \label{eq.BCD}
    \nu [\mu_1(I_1), \mu_2(I_2)] = Y_c,
\end{equation}
where $\mu_1$ and $\mu_2$ are two DNN encoders, $\nu$ is a DNN decoder. 
Under the circumstance that $I_1$ and $I_2$ exhibit homogeneity (e.g., collected by the same sensor or have similar spatial and spectral characteristics), $\mu_1$ and $\mu_2$ can be configured as siamese networks \cite{daudt2018fully}, i.e., share the same weight. This approach has been widely accepted as a paradigm for DL-based CD, as it allows effective exploitation of the temporal features.

\textbf{Techniques:} The major challenges in BCD are distinguishing semantic changes between seasonal changes and mitigating spatial misalignment as well as illumination differences. In CNN-based methods, channel-wise feature difference operations are commonly used to extract change features \cite{daudt2018fully, zhang2020feature}. Another common strategy is to leverage multiscale features to reduce the impact of redundant spatial details \cite{hou2021high}. Multiscale binary supervisions are also introduced in \cite{peng2019end} to align the embedding of change features. As an effective technique to aggregate global context, the attention mechanism is also widely used in CD of RSIs. Channel-wise attention is often used to improve the change representations \cite{li2022remote, peng2021scdnet}, while spatial attention is often used to exploit the long-range context dependencies \cite{chen2020dasnet, shi2021deeply}.

Another research focus in BCD is to model the temporal dependencies in pairs of RSIs. In \cite{chen2019change} a multilayer RNN module is adopted to learn change probabilities. Graph convolutional networks are also an efficient technique to propagate Land Cover Land Use (LCLU) information to identify changes\cite{wu2021multiscale}. Recently, Vision Transformers (ViTs) \cite{dosovitskiy2020image,li2024casformer} have emerged and gained great research interests in the RS field \cite{ding2022looking,hong2023cross}. There are two strategies to utilize ViTs for CD in RSIs. The first is to replace CNN backbones with ViTs to extract temporal features, such as ChangeFormer\cite{yuan2022transformer} and ScratchFormer\cite{noman2024remote}. Meanwhile, ViTs can also be used to model the temporal dependencies. In BiT\cite{chen2021remote}, a transformer encoder is employed to extract changes of interest, while two siamese transformer decoders are placed to refine the change maps. In CTD-Former\cite{zhang2023relation}, a cross-temporal transformer is proposed to interact between the different temporal branches.

\subsection{Multi-class CD/Semantic CD}

\textbf{Background:} In BCD, the results only indicate location of the change, leaving out the detailed change type. This is often not informative enough to support RS applications. In contrast, multi-class CD (MCD) refers to the task of classifying changes into multiple predefined classes or categories \cite{bovolo2015time}. On the other hand, semantic change detection (SCD) is introduced in recent DL-based CD literature to classify and represent the pre-event and after-event change classes \cite{daudt2019multitask, yang2021asymmetric}. Although there are slight differences in the representation of results, both MCD and SCD enable a detailed analysis of the changed regions, e.g., identifying the major changes and calculating the proportion of each type of change. The results can further be represented in an occurrence matrix indicating pre-event and after-event LCLU classes, such as presented in Fig.\ref{Fig.tasks}(b).

\textbf{Architectures:} MCD/SCD, with its provision of more detailed information, is indeed a more challenging task compared to BCD due to the need for modeling semantic information in particularly changed areas. According to the order of semantic modeling and CD, conventional methods for MCD can be roughly divided into two types, i.e. the post-classification comparison \cite{singh1990digital} and compound classification \cite{bruzzone1997iterative, wu2017post}. In DL, it is feasible to perform multi-task learning by jointly using different training objectives. There are two types of deep architectures for MCD/SCD in RSIs. The first architecture applies the common CD architecture in Equation. (\ref{eq.BCD}), and fuses bi-temporal information to classify multiple change types \cite{mou2018learning, zhu2022landuse}. The second approach employs a joint learning paradigm to learn semantic features and change representations simultaneously through different network branches \cite{daudt2019multitask, yang2021asymmetric}. This can be formulated as follows:

\begin{equation}
    \begin{aligned}
    & \phi_1 [\mu_1(I_1)] = Y_1, \phi_2 [\mu_1(I_1)] = Y_2,\\
    & \nu [\mu_1(I_1), \mu_2(I_2)] = Y_c,
\end{aligned}
\end{equation}
where $\phi_1$, $\phi_2$, and $\nu$ are three DNN modules that project the temporal features into semantic maps $Y_1, Y_2$ and change map $Y_c$, respectively.

\textbf{Techniques:} The techniques used in MCD/SCD can be categorized into two types: i) spatio-temporal fusion \cite{zhu2022landuse, zheng2022changemask} and ii) semantic dependency modeling \cite{ding2024scannet}.
In \cite{mou2018learning}, Mou et al. made an early attempt to employ DNNs for MCD. It is a joint CNN-RNN network where the CNN extracts semantic features, while the RNN models temporal dependencies to classify multi-class changes.

\subsection{Time-series CD}

\textbf{Background:} Differently from common CD studies that analyze bi-temporal RSIs, Time-Series CD (TSCD) aims to capture changes that have occurred over multiple periods or across a series of temporal images. This can better characterize the dynamics of change \cite{li2023sartscc} and discriminate between transient fluctuations and persistent changes, leading to more reliable and informative CD results. Conventional algorithms analyze the temporal curves to model the change patterns. Among these algorithms, \textit{trajectory classification} models the trajectory in change regions, \textit{statistical boundary} detects departure from common variations to detect changes, and \textit{regression} models the long-term momentum in the observed regions\cite{stahl2023automated}. Since these methods commonly do not consider spatial contexts, they are sensitive to noise and seasonal variations. Moreover, they have difficulty modeling complex or long-term change patterns. 

\textbf{Architectures:} Due to the scarcity of training data, DL-based TSCD did not emerge until very recent years. An intuitive approach is to employ RNNs to model temporal variations in time-series observations, as RNNs were originally designed for sequence processing. In \cite{yuan2020using} Long-Short-Term Memory (LSTM) network, a more delicate type of RNN is first introduced to detect and predict the burned areas in forests. Experimental results reveal that LSTM can better model the nonlinear characteristics in temporal data. In \cite{he2024timeseries} a temporal semantic segmentation method for time-series images is proposed. LSTM is employed to classify the spectral vectors into different LCLU types at different timestamps.

In these LSTM-based methods, the analysis is limited to the temporal dimension. Although the method in \cite{he2024timeseries} involves analysis of the spatial consistency, this is conducted as post-processing to reduce noise and is not end-to-end trainable. To overcome this limitation and to consider the spatial context in time-series RSIs, in \cite{sefrin2020deep} LSTM is combined with a CNN for joint spatiotemporal analysis. A CNN is employed to project time-series RSIs into spatial features, followed by an LSTM to model the temporal dependencies. This can be formulated as:

\begin{equation}
    \begin{aligned}
    & x_i = \psi (I_i),\\
    & \omega [x_1, x_2, ..., x_t] = [h_1, h_2, ..., h_t],\\
    & \nu [h_1, h_2, ..., h_t] = Y_c,
\end{aligned}
\end{equation}
where $i=1,2,...t$ is the time step, $x_i$ and $h_i$ are the extracted spatial and temporal features, $\psi$ and $\omega$ are the CNN and RNN units, respectively. $\nu$ can be a $softmax$ operation in multi-date LCLU CD applications\cite{sefrin2020deep}, or an anomaly detection function in disaster monitoring applications \cite{saha2022change}.



\section{CD with Limited Samples} \label{sc3}

\begin{figure*}[t!]
	\begin{center}
        \includegraphics[width = 0.8\textwidth]{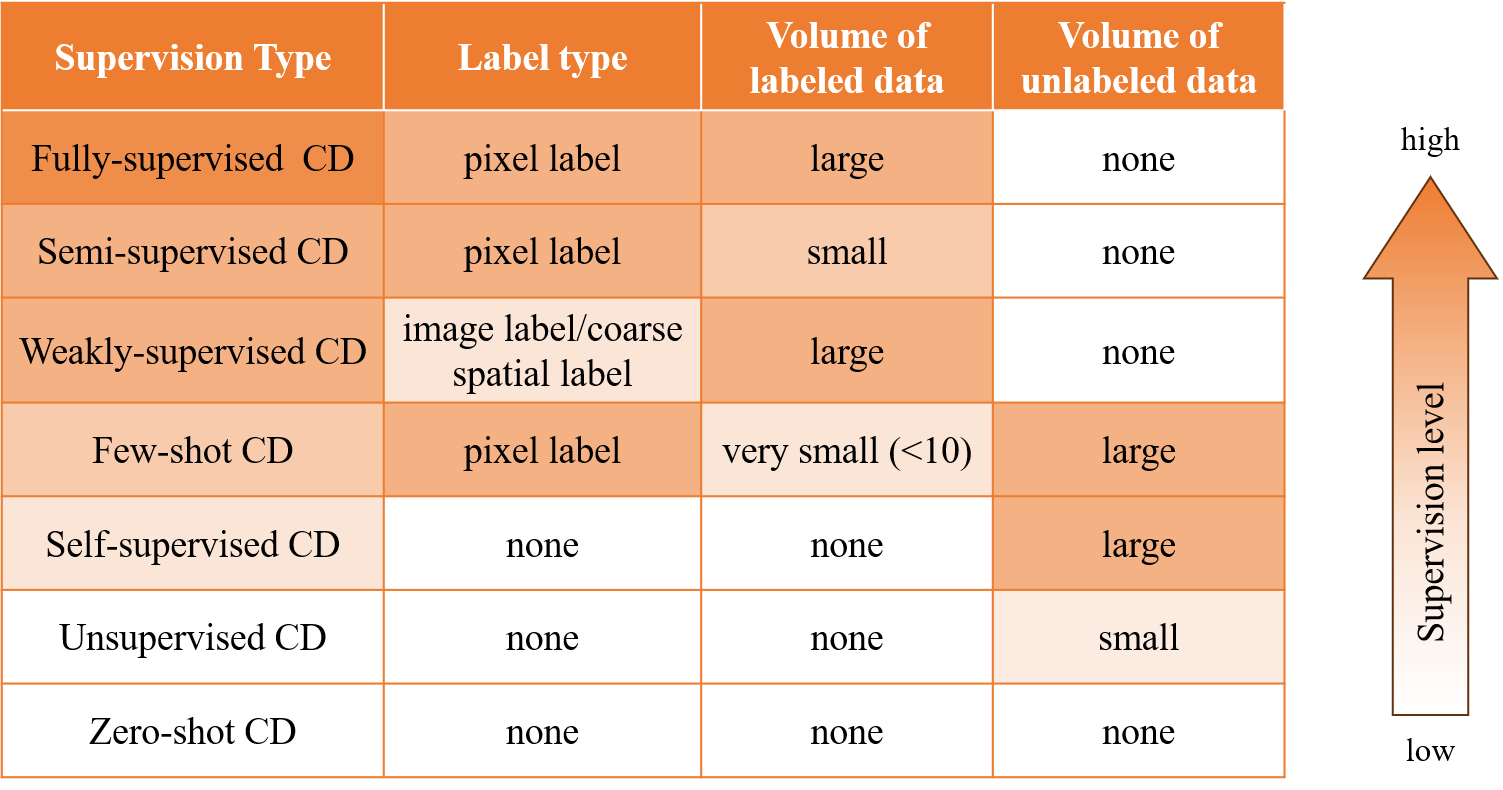}
	\end{center}
	\caption{Comparison of annotation and data volume in different CD learning paradigms.}
	\label{fig.learning_types}
\end{figure*}

To advance DL in real-world CD applications, numerous studies have been conducted on training DNNs for CD in training sample-limited experimental setups in recent years. Depending on the strength of supervision introduced in the training, sample-efficient learning of CD DNNs can be divided into 4 categories, including semi-supervised (SMCD), weakly supervised CD (WSCD), self-supervised CD (SSCD), and unsupervised CD (UCD). For readers to easily comprehend the supervision strength in different learning settings, Fig.\ref{fig.learning_types} represents the data and annotations required in each taxonomy. For simplicity, some close supervision settings are merged. The few-shot CD and zero-shot CD are incorporated into SMCD and UCD, respectively.

Furthermore, Table.\ref{Table.Strategy} summarizes various learning strategies and techniques in the literature. It is worth noting that many of these strategies can be applied to different supervision conditions. In the following, we elaborate on each supervision category and introduce the commonly used strategies, methodologies, and techniques.

\begin{table*}[htbp]
\centering
    \caption{Overview of the strategies and techniques developed to address data scarcity problem in CD.}
    \resizebox{0.9\linewidth}{!}{%
        \begin{tabular}{l|l|r}
        \toprule
            \textbf{General Strategies} & \textbf{Specific Strategies} & \textbf{Methodologies \& Techniques}  \\
            \hline
            \multirow{6}*{Auxiliary regularization} &  \multirow{2}*{Adversarial regularization} & Entropy adversarial loss\cite{peng2021semicdnet}\\
            \cline{3-3}
            & & Adversarial change masking \cite{wu2023fully}\\
            \cline{2-3}
             & \multirow{4}*{Consistency regularization} & Temporal consistency \cite{ding2022bi, hu2022hypernet}\\
            \cline{3-3}
             & & Image perturbation consistency \cite{bandara2022revisiting} \\
            \cline{3-3}
             & & Feature perturbation consistency \cite{yang2023revisiting} \\
            \cline{3-3}
             & & Perturbation consistency \& sample selection \cite{wang2024stcrnet, zuo2024robust}  \\
            \hline
            \multirow{14}*{Pseudo supervision} & \multirow{5}*{Pseudo Labeling} & Ensemble of multi-scale predictions \cite{shi2023multilayer} \\
            \cline{3-3}
             & & Ensemble of historical predictions \cite{wang2022reliable} \\
            \cline{3-3}
             &  & Ensemble of multi-model predictions \cite{yuan2024dynamically}  \\
            \cline{3-3}
             &  & Ensemble of multi-temporal predictions \cite{ding2024scannet} \cite{zheng2024detail}  \\
            \cline{3-3}
             &  & Ensemble of teacher-student predictions \cite{yang2024ecps} \cite{kondmann2023semisiroc}  \\
            \cline{2-3}
             & \multirow{5}*{Uncertainty filtering} & IoU voting \cite{yuan2024dynamically} \cite{wang2022reliable} \\
            \cline{3-3}
             &   & Entropy measure \cite{wang2022semisupervised} \cite{sun2023semibuildingchange} \\
            \cline{3-3}
             &   & Similarity measure \cite{ding2024scannet} \cite{zhang2019land} \cite{tang2022unsupervised} \\
            \cline{3-3}
             &   & Class rebalancing \cite{hou2023deep} \\
            \cline{3-3}
             &   & Contrastive sampling \cite{wang2022semisupervised} \cite{zhang2024remote} \\
            \cline{2-3}
            &  \multirow{4}*{Pre-detection supervision} & Image algebra methods \cite{chen2019change, Du2019Unsupervised}  \\ 
            \cline{3-3}
            &   & Image transformation methods \cite{Gao2016Automatic, Song2018Change} \\ 
            \cline{3-3}
            &   &  Object-based image analysis \cite{Gong2017Superpixel} \\ 
            \cline{3-3}
            &   & Saliency detection  \cite{Geng2019Saliency} \\ 
            \hline
            \multirow{7}*{Coarse-to-fine refinement} & \multirow{3}*{Change activation mapping} &  Multi-scale CAMs \cite{cao2023multiscale}\cite{lu2024weakly}\\
             \cline{3-3}
             &  & Mutual learning \cite{zhao2024pixellevel} \\
            \cline{3-3}
             &  & GradCAM++ \cite{dai2023siamese} \\
            \cline{2-3}
            & \multirow{4}*{Difference refinement} & Difference clustering \cite{kalita2021land}\\
            \cline{3-3}
            & & Guided anisotropic diffusion \cite{daudt2023weakly} \\
            \cline{3-3}
            & & CRF-RNN \cite{andermatt2021weakly} \\
            \cline{3-3}
            & & Change Masking \& Classification \cite{andermatt2021weakly} \cite{qiao2024revolutionizing} \\
            \hline
            \multirow{11}*{Representation learning} & \multirow{3}*{Graph representation} & Super-pixel graph \cite{saha2021semisupervised, wang2021dynamic, lin2023hyperspectral} \\
            \cline{3-3}
             & & Feature graph \cite{sun2022semisanet} \\
            \cline{3-3}
             & & Difference graph \cite{liu2019semisupervised, tang2022unsupervised} \\
            \cline{2-3}
             & \multirow{3}*{Contrastive learning} &  Data augmentation  \cite{feng2023detection, ou2022hyperspectral, zou2023transformer} \\
             \cline{3-3}
              & & Multiple clues  \cite{jiang2023self, huang2023contrastive, wang2023self, chen2021self, chen2022self, kuzu2024forest, yang2023multicue, qu2023tdsscd}  \\
            \cline{3-3}
              & & Pseudo label contrast \cite{9538396, adebayo2023detecting, zong2024multi}  \\
            \cline{2-3}
             & \multirow{3}*{Masked image modeling} & Large-scale MIM \& fine-tuning \cite{sun2022ringmo}\cite{cui2023hybrid} \\
            \cline{3-3}
             & & Contrastive mask image distillation \cite{muhtar2023cmid} \\
            \cline{3-3}
             & & Multi-modal MIM \cite{zhang2023self} \\
            \cline{2-3}
             & \multirow{4}*{Generative representation}  &  Autoencoder and its variants \cite{Zhange2016mapping, Liu2019Stacked, Chen2022Unsupervised} \\
            \cline{3-3}
             &  & Deep belief networks \cite{Gong2016Change, Zhao2017Discriminative} \\
            \cline{3-3}
             &  & Generative adversarial networks \cite{Lei2021Spectral, wu2023fully} \\
            \cline{3-3}
             &  & Denoising diffusion probabilistic models \cite{bandara2024ddpmcd} \\
            \hline
            \multirow{5}*{Augmentation} &  \multirow{3}*{Image augmentation} & Background-mixed augmentation \cite{huang2023backgroundmixed} \\
            \cline{3-3}
             & & Pseudo change pair generation \cite{zheng2021change}\cite{gao2024building}\\
            \cline{3-3}
             & & Patch exchange \cite{chen2023exchange}\cite{gao2024building}\\
            \cline{2-3}
             & \multirow{2}*{Change augmentation} & Object masking \& inpainting \cite{quan2023unified}\cite{seo2023selfpair} \\ %
            \cline{3-3}
             & & Change instance generation \cite{zhu2023data}\cite{zheng2023scalable}\\
            \hline
            \multirow{4}*{Leveraging external knowledge} & \multirow{2}*{Leveraging VFMs} & Fine-tuning VFMs \cite{ding2024samcd, li2024new}  \\
            \cline{3-3}
             &  & Prompt learning \cite{zheng2024segment} \\
            \cline{2-3}
             & \multirow{2}*{Transfer learning} &  Classifying VGGNet features \cite{saha2019unsupervised, Saha2022Patch} \\
            \cline{3-3}
             & ~ &  Metric learning \cite{bandara2023deep, liu2020convolutional} \\
        \bottomrule
        \end{tabular} \label{Table.Strategy} }
\end{table*}

\subsection{Semi-supervised CD}
Semi-supervised learning presupposes the availability of only a limited volume of labeled data for training. In scenarios where labeled samples are extremely scarce, this paradigm transitions into the domain of few-shot change detection. This necessitates intrinsic learning of the change patterns that can be generalized across diverse instances of change.

\textbf{Pseudo Labeling:} Pseudo labeling allows a DNN to generate pseudo labels for unlabeled data based on its predictions, thus effectively augmenting the training dataset. In segmentation-related tasks, pseudo labels can be obtained by thresholding the predictions of DNNs.

Since single DNN predictions may contain many errors, various methods combine multiple predictions to enhance the robustness of pseudo-labeling. In \cite{shi2023multilayer} pseudo labels are obtained by composing and voting multi-scale predictions. In \cite{wang2022reliable}, historical models are used during training to produce ensemble predictions. By calculating the mean Intersection over Union (IoU) in historical predictions, the reliable results are selected as pseudo labels to train the unlabeled data. The method in \cite{yuan2024dynamically} utilizes multiple DNNs to produce multiple predictions and also performs IoU calculations to generate reliable labels. In \cite{ding2024scannet} a cross-temporal pseudo-labeling technique is introduced. The semantic similarity between multitemporal predictions is calculated to select the high-confident pixels. In \cite{yang2024ecps}, a sophisticated cross-pseudo supervision method is proposed within the Teacher-Student (TS) learning paradigm. The knowledge learned in a teacher model is distilled to supervise the student models, and the predictions of multiple student models are composed to generate reliable pseudo labels. Kondmann et al. \cite{kondmann2023semisiroc} employ an unsupervised method as the teacher model, subsequently train and fine-tune different CD models with pseudo labels from the teacher model. In \cite{zhan2023s}, the method employs superpixel segmentation to create objects and enable self-supervised learning through object overlaps in bitemporal images. It produces and integrates multiscale object-level and pixel-level difference images and utilizes temporal prediction for SSCD.

The essence of pseudo-labeling is minimizing the errors and uncertainty in generated labels while enhancing guidance for critical cases. Therefore, it is important to measure the certainty of DNN predictions. If the pseudo labels are generated by multiple methods, the number of votes can be deemed the confidence score \cite{kondmann2023semisiroc}. In \cite{yuan2024dynamically} and \cite{wang2022reliable} the certainty is measured through IoU in multiple predictions. For a single DNN prediction on unlabeled data, low entropy indicates high confidence, and entropy-based objectives are commonly used to filter uncertain predictions \cite{wang2022semisupervised} \cite{sun2023semibuildingchange}. In \cite{zhang2019land} similarity measures and uncertainty calculations are combined to map the pseudo CD labels. To improve the guidance for minor classes (i.e., changes), Hou et al. \cite{hou2023deep} cluster the extracted deep features to generate pseudo labels and rebalance the \textit{change/non-change} instances in pseudo labels to strengthen the learning of minority class (i.e., \textit{changes}). Furthermore, uncertain predictions also contain potential knowledge. In a contrastive learning paradigm, reliable and unreliable pixels can be sampled as positive and negative samples, thus improving the representation of temporal semantic features \cite{wang2022semisupervised, zhang2024remote}.

\textbf{Auxiliary regularization:}
To facilitate training on unlabeled data, a common strategy is to introduce auxiliary training objectives or regularization. This can constrict the optimization landscape and regularize DNNs to learn noise-resistant change representations. In \cite{ding2022bi} Ding et al. propose a temporal similarity regularization to optimize learning of temporal semantics in SCD. This objective drives DNNs to embed similar features in unchanged areas and different semantics in changed areas. In \cite{ding2024samcd} it is extended with temperature regularization to model the implicit semantic latent in the BCD. In \cite{zheng2024detail} temporal regularization is implemented in the form of mutual supervision with pseudo labels.  In \cite{hu2022hypernet} a focal cosine loss is designed to align feature representations in unchanged areas for SSCD of hyperspectral images. It assigns greater weights to hard positive samples to emphasize the learning of critical samples. 

In \cite{peng2021semicdnet} adversarial learning is introduced to align the feature distributions of unlabeled data with the labeled data, thus promoting GT-like results. In \cite{dong2020self}, adversarial learning is introduced to learn consistent feature representations in bitemporal images. The CD results are then derived by clustering the different features.

\begin{figure}[t!]
	\begin{center}
        \includegraphics[width = 0.5\textwidth]{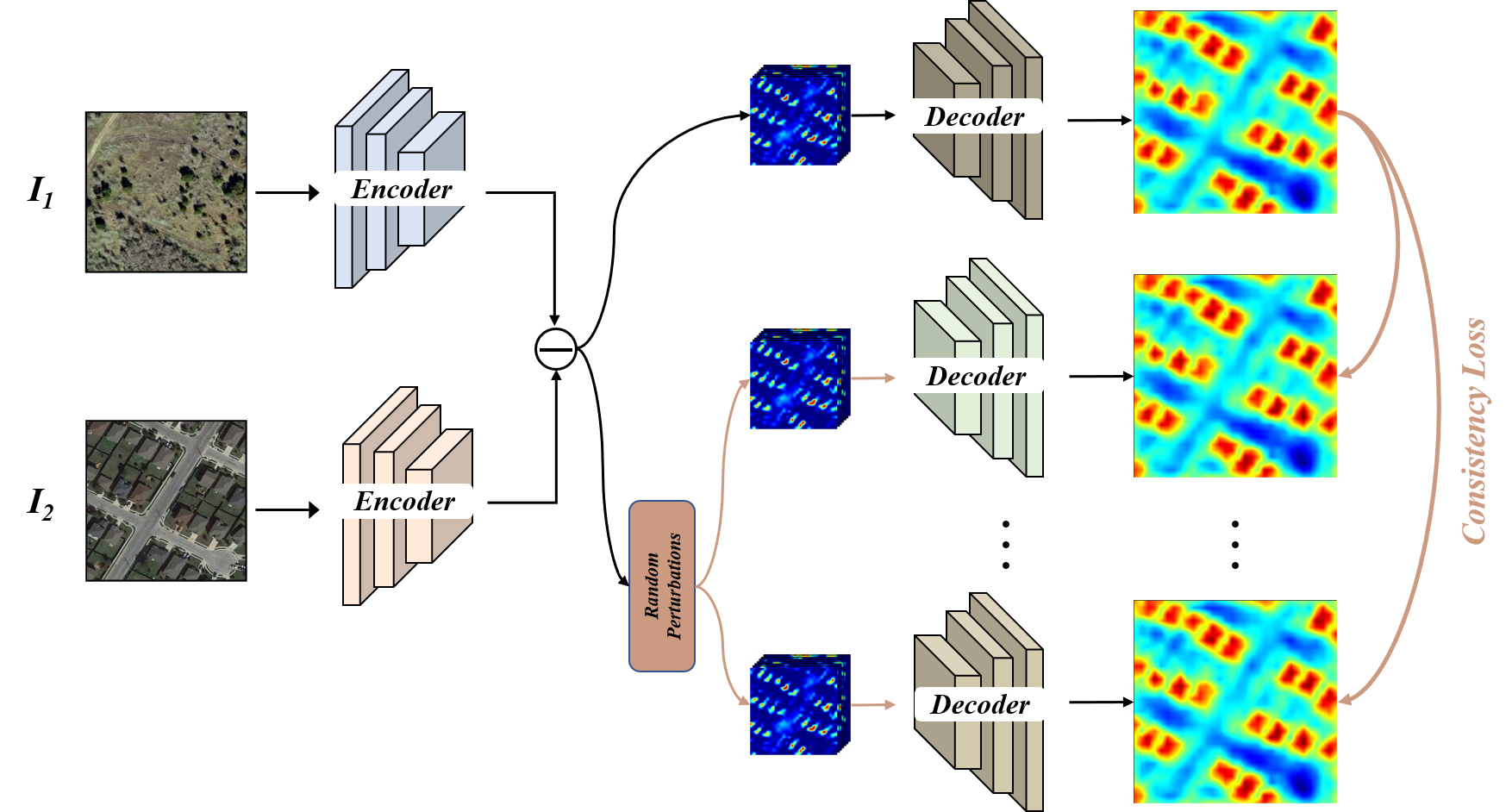}
	\end{center}
	\caption{Consistency regularization for WSCD \cite{bandara2022revisiting}. Random perturbations are applied to the change representations, and a consistency loss is calculated between the origninal and perturbed CD results to improve the robustness of CD models.}\label{fig.CR}
\end{figure}

Among auxiliary regularization-based approaches, Consistency Regularization (CR) is an effective strategy to enhance the model generalization. CR applies spatial or spectral perturbations to unlabeled data, training the model to reduce discrepancies between varying perturbations of the same image \cite{sohn2020fixmatch}.

Bandara et al. \cite{bandara2022revisiting} first introduce CR to WSCD, and extend perturbations from images to feature differences. A paradigm for CR is proposed in the context of WSCD, which involves different types of perturbations, such as random feature noising, random feature drop, feature cutout, and instance masking. Similarly, Yang et al. \cite{yang2023revisiting} extend the CR paradigm with dual stream feature-level perturbations, which greatly improves the generalization even with a very small proportion of training samples. A simplified paradigm of this CR learning under a teacher-student knowledge distillation framework is illustrated in Fig.\ref{fig.CR}.

Building on top of the CR paradigm, many literature methods investigate to improve WSCD through advanced DNN designs and sample selection mechanisms. In \cite{zhang2023joint}, rotation augmentation is introduced in CR-based WSCD, and class-wise uncertainties are calculated to alleviate the class imbalance issue. Wang et al. \cite{wang2024stcrnet} introduce a reliable sample selection mechanism that selects samples with stable historical predictions during training. In \cite{han2024c2fsemicd}, a coarse-to-fine CD network with multiscale attention designs is designed as the backbone for CR-based WSCD. In \cite{zuo2024robust} selection, trimming and merging of reliable instances is performed to enhance the robustness of extracted change instances. Hafner et al. consider multi-modal data as different views of the same regions and employ CR across different modalities to learn robust built-up changes \cite{hafner2023semisupervised}.

\textbf{Graph Representation:} Graph neural networks (GNNs) are a family of DNNs that are adept at modeling relationships. Since GNNs can be trained with partial labels, they are well suited to semi-supervised learning settings \cite{liu2019semisupervised, saha2021semisupervised}. A crucial step in graph learning is graph construction. The literature methods can be categorized into superpixel-based \cite{saha2021semisupervised, wang2021dynamic, lin2023hyperspectral}, feature-based \cite{sun2022semisanet}, and difference-based \cite{liu2019semisupervised, tang2022unsupervised} graph construction. 

Liu et al. \cite{liu2019semisupervised} first introduced graph learning in the context of SMCD. The differences between temporal features are calculated to construct change graphs, while adversarial learning is also introduced to train the graphs constructed with unlabeled data. Saha et al. construct change graphs with multi-temporal parcels, and propagate change information from labeled parcels to unlabeled ones through training iterations \cite{saha2021semisupervised}. Tang et al. \cite{tang2022unsupervised} employ a multi-scale Graph Convolutional Network (GCN) to capture long-range change context and generate pseudo labels with similarity metrics. In \cite{wang2021dynamic} a method for dynamic graph construction in SAR image CD is presented. It constructs graphs from three-channel pixel blocks and dynamically updates graph edges based on trained features. The method in \cite{lin2023hyperspectral} combines superpixel graph modeling and pixel-level CNN embedding for SMCD in hyperspectral images. It introduces a graph attention network (GAT) to capture temporal-spatial correlations via an affinity matrix and uses CNN layers to merge features to map changes. In \cite{sun2022semisanet}, GAT is incorporated into a CR learning framework to learn robust multi-temporal graph representations. In \cite{kalinicheva2020unsupervised} graph is employed to represent and cluster the change evolutions for unsupervised TSCD.

\subsection{Weakly supervised CD}
While CD is a fine-grained segmentation task that requires pixel-level annotations, in the weakly supervised learning setting, only coarse-grained labels such as points, surrounding boxes, scribbles, and image categories are available. WSCD enables easy construction of a CD training set, as it does not require intensive human annotation. However, it does not mitigate the scarcity of change samples. 

Most of the WSCD methods utilize image-level labels. The labels indicate either the image categories \cite{cao2023multiscale} or the image pair (\textit{change/nonchange} \cite{dai2023siamese}). Meanwhile, various types of coarse CD labels are also utilized in literature studies, including point labels \cite{fang2023point}, low-resolution labels \cite{zheng2021weakly}, patch-wise labels \cite{qiao2024revolutionizing} and box labels \cite{khan2017forest}. The differences in these supervisions derive different methodologies of utilizing and recovering spatial information. Two major categories of WSCD methodologies that correspond to image-level supervision and coarse CD supervision are change activation mapping and difference refinement, respectively.

\begin{figure}[t!]
	\begin{center}
        \includegraphics[width = 0.5\textwidth]{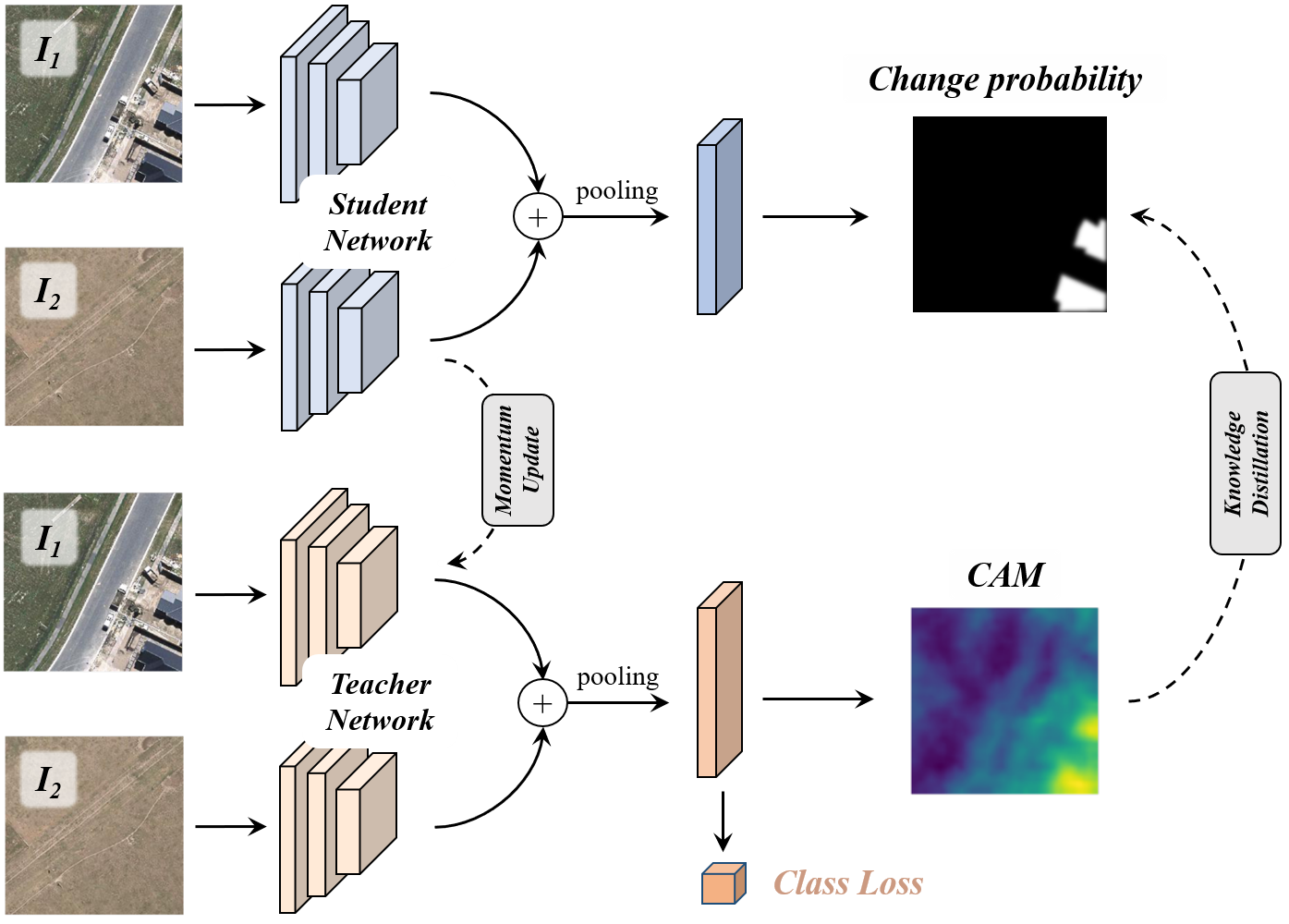}
	\end{center}
	\caption{Refining CAM for SMCD within a teacher-student framework \cite{lu2024weakly}. A CAM is obtained with image-level supervision (class loss), and is refined through knowledge distillation.} \label{fig.CAM_TS}
\end{figure}

\textbf{Change activation Mapping:} 
This strategy is frequently employed in WSCD to parse image-level label into spatial change representations. First, an image encoder is trained with image-level information, then the feature responses in the late layers, i.e., class activation maps (CAMs), are utilized to generate coarse pseudo labels. However, CAMs contain only coarse feature responses and do not indicate fine-grained change details. To improve the accuracy and robustness of CAMs, Cao et al. \cite{cao2023multiscale} ensemble multi-scale CAMs and propose a noise correction strategy to generate reliable pseudo labels. The method in \cite{lu2024weakly} also adopts a multi-scale approach. It extracts more robust and accurate change probability maps through knowledge distillation and multi-scale sigmoid inference, as illustrated in Fig.\ref{fig.CAM_TS}. The method in \cite{zhao2024pixellevel} introduces mutual learning between different time phases. It utilizes CAMs derived from the original image and the affine transformed image to improve the certainty of change mapping and incorporates contrastive learning to enlarge the distance between changed representations and unchanged representations. In \cite{dai2023siamese} GradCAM++ is introduced to weight the multi-scale CAMs. It also leverages multi-scale and transformation consistency regularization to improve the quality of CAMs.

\textbf{Difference Refinement:} In comparison to image-level labels, coarse CD labels contain a certain degree of spatial information and thus can be utilized to train a coarse CD model. After mapping the differences, various kinds of techniques are developed to refine and highlight the salient change regions.

Several methods employ conventional machine learning methods to perform the refinement. In \cite{zheng2021weakly}, the refinement is achieved through bitemporal comparison and morphological filtering operations. In \cite{khan2017forest}, a candidate suppression algorithm is designed to reduce the overlapping box candidates and select the most confident candidate regions that indicate changes. In \cite{kalita2021land} temporal features are extracted by contrastive learning, and the mapping from difference image to CD result is achieved through PCA and K-Means algorithms. 

In contrast to refinement on the CD results, several methods refine the labels to perform fine-grained supervision. The method in \cite{lu2020weakly} first calculates a difference map through edge mapping and superpixel segmentation algorithms, then trains a denoising autoencoder to refine the pre-classification results. Fang et al. apply region growth on point labels and DNN predictions to expand the annotations and propose a consistency alignment objective to align the coarse and fine predictions \cite{fang2023point}. In \cite{daudt2023weakly}, the training of a CD CNN and the refinement of the results are carried out iteratively to reduce the errors in the noisy crowd-sourced labels. A guided anisotropic diffusion algorithm is introduced to filter the wrong predictions while preserving the edges.

\begin{figure*}[t!]
	\begin{center}
        \includegraphics[width = 0.8\textwidth]{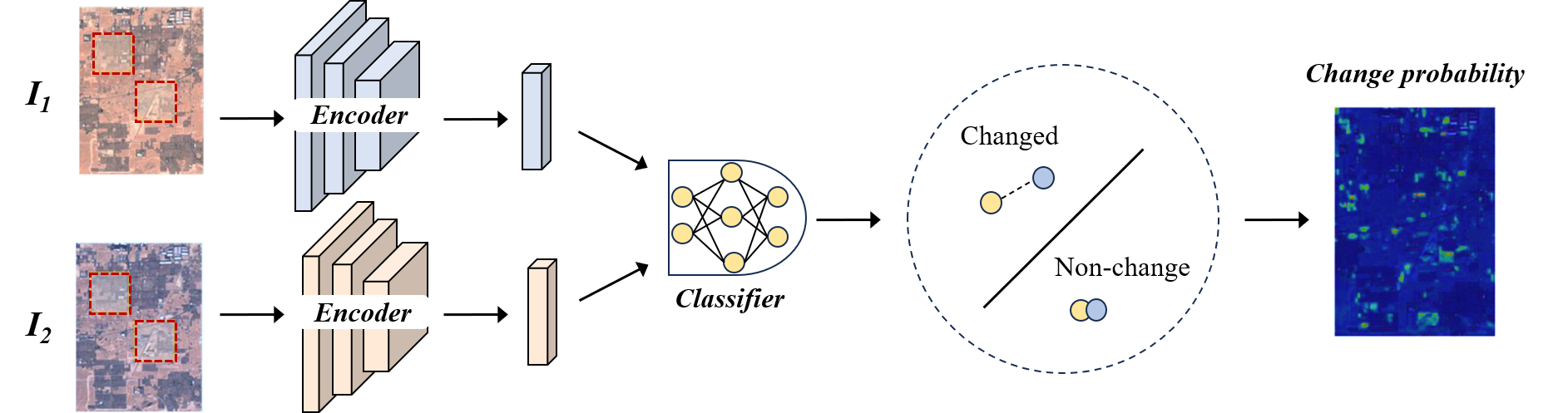}
	\end{center}
	\caption{A simplified paradigm of contrastive learning for SSCD \cite{chen2022self}. Croppsed RSIs in the same and different locations construct positive and negative change pairs.}
	\label{fig.Contrastive}
\end{figure*}

Differently from these approaches, the method in \cite{andermatt2021weakly} utilizes object-level class labels to perform WSCD. It first compares image pairs with a Siamese Unet and then masks the changed object to classify its category. To enable accurate masking of the changed object, a CRF-RNN (Conditional Random Fields as Recurrent Neural Network) layer is employed to integrate spatial details from the original image. Similarly to this object-masking approach, the method in \cite{qiao2024revolutionizing} masks and re-segments superpixels as interesting instances (buildings), and utilizes a voting mechanism to classify the changed instances (damaged buildings).


\subsection{Self-supervised CD} 

Self-supervised learning exploits the inherent consistency within data to learn sensor-invariant and noise-resilient semantic representations. Leveraging the capability of self-supervised learning, SSCD learns to discriminate temporal variations in unlabeled RSIs. It is worth noting that SSCD can be regarded as a distinct subclass within the broader category of UCD, but they typically require extensive pre-training in the target domain. Additionally, many approaches employ SSCD for pretraining and still require fine-tuning on target datasets.

\textbf{Contrastive Learning:} This strategy constructs and compares positive and negative pairs to exploit the structure and relationships within unlabeled data. In CD, bi-temporal images are often utilized to construct the contrastive pairs. By maximizing the consistency among positive pairs and the difference among negative pairs through contrastive losses, DNNs are trained to exploit feature embeddings that can capture temporal similarities and discrepancies. Fig.\ref{fig.Contrastive} illustrates a simplified paradigm of contrastive learning, where change pairs are constructed with cropped RSIs at the same and different locations. The mapping of pre-trained representations into CD results further categorizes two major types of methods: fine-tuning-based and thresholding-based.

\par The fine-tuning-based methods use CD labels to retrain based on the pre-trained model obtained from self-supervised methods. Common methods utilize data augmentation methods for comparative learning. The results of data augmentation based on the same sample are regarded as positive samples. Feng et al. \cite{feng2023detection} obtain a pre-trained model based on SimSiam and unlabeled samples. Multiple data augmentation methods are often combined to generate positive samples, and then the pre-trained model is directly fine-tuned \cite{ou2022hyperspectral} \cite{zou2023transformer}. In addition to data augmentation, some studies construct contrastive learning by mining multiple clues, such as multi-level contrast and multi-feature contrast. Jiang et al. \cite{jiang2023self} design global-local contrastive learning, where global and local contrastive learning respectively implement instance-level and pixel-level discrimination tasks. Huang et al. \cite{huang2023contrastive}  propose a soft contrastive loss function to improve the inadequate feature tolerance. In the downstream CD fine-tuning task, the features of different receptive fields are captured by a multiscale feature fusion module and combined with a two-domain residual attention block to obtain long-range dependencies on spectral and spatial dimensions. The method in \cite{wang2023self} proposes a multilevel and multi-granularity feature extraction method and applies contrastive learning to obtain the pretrained model. A multilevel CD is performed by fine-tuning the network with limited samples.

\par The thresholding-based methods derive the CD map from dual feature maps using thresholding, thus no labeled samples are used for fine-tuning. Contrastive learning based on multiple clues has also been used in these methods. The method in \cite{chen2021self} pretrains the model using a pseudo-siamese network and multiview images and then generates binary CD maps through feature distance measurement and thresholding. In \cite{chen2022self}, shifted RSI pairs are leveraged to train pseudo-siamese networks, performing pixel-level contrastive learning. Kuzu et al. \cite{kuzu2024forest} employ instance-level (BYOL, SimSiam) and pixel-level (PixPro, PixContrast) methods to derive pre-trained models and directly produce CD maps using DCVA. In \cite{yang2023multicue}, a multicue contrastive self-supervised learning framework is designed. Beyond mere data augmentation, this approach also constructs positive sample pairs from semantically similar local patches and temporally aligned patches. The preliminary change embeddings are then obtained from the affinity matrix. The method in \cite{qu2023tdsscd} first performs contrastive learning on bitemporal RSIs, and then performs contrastive learning on early fusion and late fusion features. 
Meanwhile, pseudo label contrast has also been widely explored, which regards samples with the same class as positive pairs and samples with different classes as negative pairs. Saha et al. \cite{9538396} employ deep clustering and contrastive learning for self-supervised pre-training. Adebayo \cite{adebayo2023detecting} trains a classifier using land cover labels of available years to identify unchanged regions through post-classification comparisons.  The pre-trained model is obtained through the BYOL method based on trusted unchanged regions. He et al. \cite{zong2024multi} employ clustering to obtain pseudo labels (\textit{non-changed, changed,} and \textit{uncertain}). Furthermore, this framework introduces a self-supervised triple loss, including \textit{changed} and\textit{ non-changed} losses based on contrastive learning and an uncertain loss based on image reconstruction.

\textbf{Masked Image Modeling:} Masked Image Modeling (MIM) is a self-supervised reconstructive approach aims at learning generalized representations from extensive volumes of unlabeled data. Within the MIM paradigm, DNNs are trained to reconstruct masked image pixels or patches based on available unmasked image content. However, MIM does not provide task-specific feature representations and typically requires subsequent fine-tuning.

With large-scale pretraining using MIM, Sun et al. \cite{sun2022ringmo} constructed a foundational model for RS scenes and proved its improvements to BCD. Cui et al. \cite{cui2023hybrid} pre-train a network using multi-scale MIM and fine-tune it with labeled data. The model first processes images with convolutional structures and then extracts global information using transformers. The method in \cite{muhtar2023cmid} combines contrastive learning and MIM in a self-distillation way, allowing effective representations with global semantic separability and local spatial perceptibility. Zhang et al. \cite{zhang2023self} propose a multi-modal pretraining framework. The DNNs learn visual representations through MIM, and align them with multi-modal data through contrastive learning. A temporal fusion transformer is also proposed to transfer the pre-trained model to CD.

\begin{figure}[t!]
	\begin{center}
        \includegraphics[width = 0.5\textwidth]{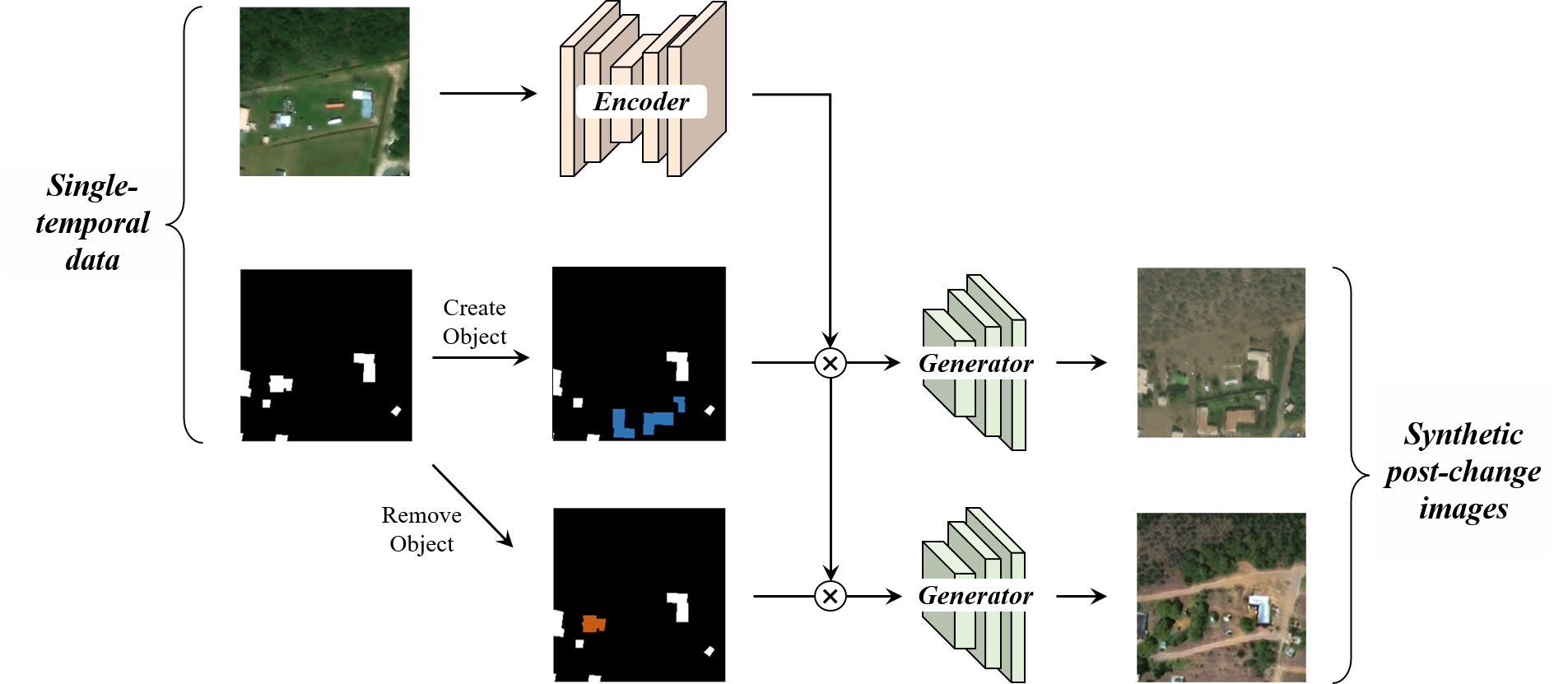}
	\end{center}
	\caption{The paradigm of semantic change augmentation in \cite{zheng2023scalable}. Post-change RSIs are synthesized with single-temporal images and instance labels.}
	\label{fig.ChangeAug}
\end{figure}

\textbf{Augmentation:} Natural changes are infrequent and registered bitemporal RSIs are difficult to collect. To overcome these limitations, in \cite{huang2023backgroundmixed} a background augmentation method is introduced for image-level WSCD. It augments samples under the guidance of background-exchanged images, enabling the model to learn intricate environmental variations. 

Several literature studies resort to augmenting semantic changes with single-temporal RSIs in segmentation datasets. In \cite{zheng2021change}, pseudo change pairs are constructed by randomly sampling labeled RSIs and mixing their semantic labels. This pseudo supervision is proved to generalize well on CD datasets without fine-tuning. In \cite{chen2023exchange}, Chen et al. propose a simple image patch exchange method to generate pseudo-multi-temporal images and pseudo labels from a wide range of single-temporal HR RSIs, facilitating the training of CD DNNs in a self-supervised manner. In \cite{gao2024building}, patches from other images are cut and pasted to create a pseudo-post-change image.

There are also several literature studies aiming to generate more diverse and realistic change pairs with instance-level augmentations. They commonly utilize instance labels in segmentation datasets to perform the creation or removal of synthetic changes, as illustrated in Fig.\ref{fig.ChangeAug}. For example, Seo et al. \cite{seo2023selfpair} implement copy-pasting or removal-inpainting operations based on the labels of ground objects. Zheng et al. \cite{zheng2023scalable} first synthesize changes by copying or removing objects, then simulate temporal variations using a GAN. Zhu et al. \cite{zhu2023data} generate object segments with a GAN and employ Poisson blending to fuse them into background images. The resulting approach enables a few-shot CD in forest scenes. Quan et al. \cite{quan2023unified} generate pseudo-change pairs by masking the instances in labeled building segmentation datasets. After pretraining on these synthesized datasets, high accuracy is yielded with few amount of labeled data for fine-tuning.

\subsection{Unsupervised CD}

\par UCD eliminates the necessity for prior training, allowing direct deployment of CD algorithms on unlabeled data. This significantly broadens the application scope of DL-based CD, representing a critical objective in the advancement of CD methodologies. However, unsupervised CD presents significant challenges for DL-based frameworks, as the training of DNNs requires task-specific objectives. To address the absence of explicit supervision in UCD, the literature identifies three principal strategies: generative representation, pre-detection supervision, and leveraging external knowledge.

\textbf{Generative Representation:} This approach uses generative models to extract features, eliminating the need for manually labeled data \cite{hong2024multimodal}.

The model extracts feature maps from the original multi-temporal image for pixel-wise comparison to generate a difference map. A distance metric, such as the Euclidean distance, combined with a threshold segmentation algorithm, derives the final CD results. Prevalent deep generative models include auto-encoders (AE), deep belief networks (DBN), generative adversarial networks (GAN) \cite{Goodfellow2014Generative}, and denoising diffusion probabilistic models (DDPM) \cite{Jonathan2020Denoising}.

\par AEs are unsupervised learning models optimized by minimizing reconstruction errors. However, vanilla AEs tend to learn redundant information (e.g., simply replicating the input data) and encounter difficulties in deriving meaningful representations within a single-layer architecture. Consequently, various variants such as stacked AE (SAE), sparse AE, denoising AE (DAE), and variational AE (VAE) have been adapted for CD tasks.

In \cite{Chen2019Fast}, an SAE-based algorithm for CD of HR RSIs employs a sparse representative sample selection strategy to reduce time complexity. Liu et al. \cite{Liu2019Stacked} use an SAE with Fisher's discriminant criterion for high-resolution SAR image CD to better distinguish between changed and unchanged features. In \cite{Hu2021Hyperspectral}, SAE served as a predictor of hyperspectral anomaly CD. Touati \cite{Touati2020Anomaly} designed a multimodal CD (MMCD) framework based on anomaly detection, noting that changed regions often exhibit significant reconstruction losses in sparse AE. Lv et al. \cite{Lv2018Deep} used a contractive AE to minimize noise and extract deep features from superpixels for the SAR image CD. In \cite{Hu2021Hyperspectral}, SAE served as a predictor of hyperspectral anomaly CD. Touati \cite{Touati2020Anomaly} designed an MMCD framework based on anomaly detection, noting that changed regions often exhibit significant reconstruction losses in a sparse AE. Lv et al. \cite{Lv2018Deep} employ a contractive AE to minimize noise and extract deep features from superpixels for the SAR image CD. In \cite{Zheng2022Unsupervised}, a cross-resolution difference learning method involving two coupled AEs was developed for CD across images of varying resolutions.

Since DAEs help reduce the impact of noise on original images, they are widely used in SAR and MMCD \cite{Zhange2016mapping, LiuGQZ18, Zhan2018Iterative, Zhan2018Log}. To mitigate the loss of spatial contextual information typically associated with vectoring operations in conventional AEs, convolutional layers have been incorporated into AEs, resulting in the development of convolutional AEs (CAEs) for CD. Bergamasco et al. \cite{Bergamasco2022Unsupervised} develop a CAE to learn multi-level difference features for multispectral CD. Wu et al. \cite{Wu2022Commonality} add a commonality constraint to CAE for MMCD applications. Furthermore, to address spatial information loss in fully connected AEs, Wu et al. \cite{wu2021unsupervised} propose a kernel principal component analysis (KPCA) convolution feature extraction model. A deep KPCA convolutional mapping network is designed following the layer-wise greedy training approach of SAE for both BCD and MCD in HR RSIs. Chen et al. \cite{Chen2022Unsupervised} present a graph-based framework to model structural relationships for unsupervised multimodal CD. It employs dual-graph convolutional autoencoders to discern modality-agnostic nodes and edges within multimodal images.

\par DBNs are another type of classic unsupervised deep model with multiple layers of restricted Boltzmann machines (RBMs). Like SAE, DBNs are trained using a layer-wise greedy approach, enabling them to extract informative features from input images. Despite their potential, DBNs have seen relatively limited application in CD. Gong et al. \cite{Gong2016Change} utilized pre-trained DBN weights as initial weights for a DNN to perform CD on SAR images. Zhao et al. \cite{Zhao2017Discriminative} designed a DNN composed of two symmetric DBNs to learn the modality-invariant features for MMCD. Jia et al. \cite{JiaChange2021} introduced a generalized Gamma DBN to learn features from different images, and Zhang et al. \cite{Zhang2016Feature} compressed features extracted by DBN into a 2D polar domain for MCD on multispectral images.

\par As a prominent framework for approaching generative AI, GANs have also been widely applied in unsupervised CD. Lei et al. \cite{Lei2021Spectral} apply GANs to learn representative features from hyperspectral images, achieving robust CD results. Saha et al. \cite{Saha2021Unsupervised} develop a GAN-based method to learn deep change hypervectors for CD on multispectral images. Ren et al. \cite{Ren2021Unsupervised} developed a GAN-based CD framework to mitigate the issues caused by unregistered objects in paired RSI. Wu et al. \cite{wu2023fully} propose an end-to-end unsupervised CD framework, jointly training a segmentor and a GAN with \textit{L1} constraints. Noh et al. \cite{noh2022unsupervised} employ GANs for image reconstruction using single temporal images in training and bitemporal images in inference, identifying changed regions by high reconstruction losses. GANs demonstrate exceptional efficacy in MMCD owing to their advanced capabilities in image style transfer. One of the major types of unsupervised MMCD, modality translation methods, predominantly leverages GANs. For instance, Niu et al. \cite{Niu2019Conditional} use a conditional GAN for modality translation between SAR and optical images, obtaining CD results through direct comparison of transformed images. Subsequent advances include sophisticated GAN architectures and training techniques for improved detection accuracy, such as cycle-consistent GAN \cite{Luppino2022Deep,Liu2022Unsupervised}, CutMix \cite{Anamaria2022Generative}, feature space alignment \cite{Luppino2024Code}, and robust fusion-based CD strategies \cite{WangCD2024}. These approaches often incorporate pre-detection techniques to isolate changed regions for more stable modality translation results, aligning with the concepts we will discuss in the following subsection.

\par DDPMs, drawing inspiration from the principles of non-equilibrium thermodynamics, have garnered significant attention in generative artificial intelligence \cite{Jonathan2020Denoising}. These models involve a diffusion process that gradually introduces random noise into the data, followed by a reverse diffusion process to reconstruct the desired data distribution from the noise. Training by reconstructing inputs makes DDPMs naturally suitable for feature extraction in CD tasks. Bandara et al. \cite{bandara2024ddpmcd} first introduced DDPMs for CD. However, subsequent works focus mainly on fully supervised CD (FSCD) \cite{Wen2024GCD}, while studies on UCD with DDPMs remain rare.

\begin{figure*}[t!]
	\begin{center}
        \includegraphics[width = 0.8\textwidth]{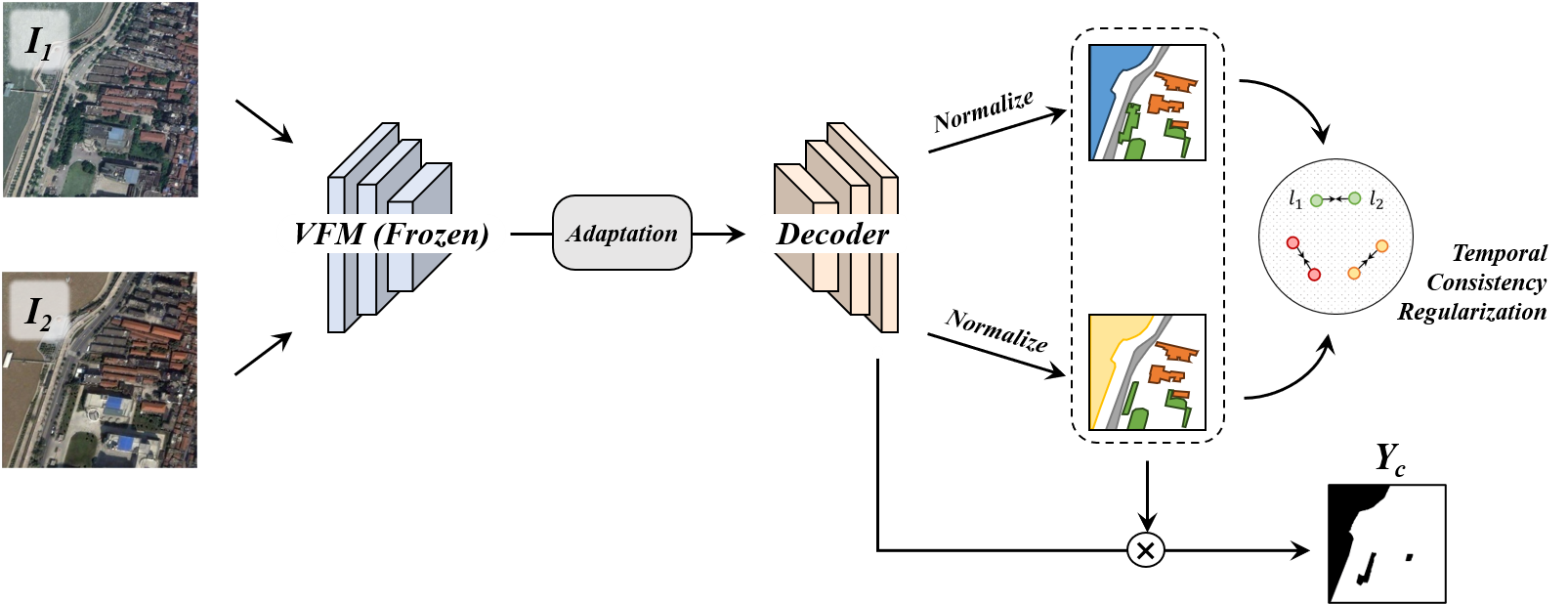}
	\end{center}
	\caption{The paradigm of leveraging VFM for CD in \cite{ding2024samcd}. VFM parameters are 'frozen' (i.e., not updated), whereas other network modules are trainable to adapt VFMs to the RS domains.}
	\label{fig.VFM_CD}
\end{figure*}

\textbf{Pre-detection Supervision:}\label{sec:pre_detection}
Although unsupervised generative models do not require labeled data to extract features from images for CD, the lack of objectives during the feature extraction process may result in suboptimal and less informative features. Additionally, the absence of labeled data can limit the learning of more advanced DL models. To address these issues, pre-detection-based approaches first generate pseudo labels using traditional unsupervised CD algorithms, then train deep CD models with the pseudo labels. This strategy emulates supervised learning paradigms for training purposes while remaining

entirely unsupervised, as it does not depend on any pre-existing labeled data. Several early DL-based CD methods have adopted this strategy.

\par The effectiveness of pre-detection supervision depends on the accuracy of pre-detection algorithms. Thus, it is crucial to design or select algorithms that suit the characteristics of input images. Synthetic Aperture Radar (SAR) images, in particular, have been extensively studied due to their unique speckle noise. Gao et al. \cite{Gao2016Automatic} developed an automatic CD algorithm using PCANet \cite{Chan2015PCANet}, which employs a Gabor wavelet transform and Fuzzy C-means clustering (FCM) to select the most reliable changed and unchanged samples from SAR images. These samples are then used to train the PCANet. Similarly, Gong et al. \cite{Gong2016Change} proposed a deep neural network-based CD algorithm for SAR images that incorporates a pre-detection algorithm based on FCM to select the most representative samples. In another study, Gong et al. \cite{Gong2017Feature} introduced an unsupervised ternary CD algorithm where deep feature representations are learned from the difference image using an SAE, effectively suppressing image noise. Geng et al. \cite{Geng2019Saliency} integrated saliency detection into CD for SAR images by designing a pre-detection algorithm to select representative and reliable samples for training the deep network. Additionally, Yang et al. \cite{yang2019transferred} combined the concept of transfer learning with pre-detection methods to broaden the application scope of CD in SAR images. Liu et al. \cite{Liu2019Local} proposed a locally restricted CNN that adds spatial constraints to the output layer, effectively reducing noise in Polarimetric SAR (PolSAR) images. This model was also supported by a pre-detection algorithm based on the statistical properties of PolSAR images. 

\par Methods tailored for multispectral, hyperspectral, and high-resolution images have also been developed. Gong et al. \cite{gong2017generative, 2019GenerativeGong} leveraged the initial difference image generated by the CVA to provide a priori knowledge for sampling training data for GANs. Shi et al. \cite{Shi2022Unsupervised} extended this approach to MCD. Du et al. \cite{Du2019Unsupervised} introduced a deep slow feature analysis (DSFA) model combined with a deep neural network to learn nonlinear features and emphasize changes. The authors employed a CVA-based pre-detection method to select samples from multispectral images for training the network. Song et al. \cite{Song2018Change} utilized PCA and image element unmixing algorithms to select training samples for a recurrent 3D fully convolutional network for binary and multiclass CD. In \cite{Hu2023Binary}, pseudo-labels from BCD were employed to guide hyperspectral MCD. For high-resolution images, pre-detection algorithms need to focus more on the spatial information within the image. Gong et al. \cite{Gong2017Superpixel} developed a high-resolution CD algorithm based on superpixel segmentation and deep difference representation. This method achieved varying pre-detection results based on different superpixel features and implemented a voting mechanism to select reliable training samples from these results. Xu et al. \cite{Xv2019Combining} used SFA as a pre-detection algorithm to select reliable samples to train a stacked DAE for high resolution RSI CD.

\textbf{Leveraging external knowledge:}
DNNs pre-trained on natural images are adept at extracting general visual features, which can be highly beneficial for the recognition tasks of RSIs. An early exploration by Saha et al. \cite{saha2019unsupervised} utilized a CNN encoder pre-trained on natural optical images to extract bitemporal features, which were then pixel-wise compared to classify changes. Subsequently, Saha et al. \cite{Saha2022Patch} applied the pre-trained VGG network as a feature extractor for planetary CD. Bandara et al. \cite{bandara2023deep} introduce multiple bitemporal constraints based on metric learning to transfer the inherent knowledge from pre-trained VGG networks to the RS target domain. The approach in \cite{liu2020convolutional} initially transfers deep features pre-trained on semantic segmentation datasets, then fine-tunes them with distance constraints and pseudo-change labels to enhance relevance. Furthermore, in \cite{Zhan2022Transfer}, object-based image analysis was leveraged to refine feature extraction with a pre-trained CNN. To better tailor the features for the RS domain, a clustering function based on feature distance calculation was introduced in \cite{sublime2019automatic}. Yan et al. utilize multi-temporal remote sensing indices as domain knowledge to guide the contrastive learning of change representation \cite{yan2023domain}. 

Recently, Visual Foundation Models (VFMs) such as CLIP \cite{radford2021CLIP} and Segment Anything Model (SAM) \cite{Kirillov2023Segment} have emerged and gained significant research interest. VFMs, pre-trained on web-scale datasets, are designed to capture universal feature representations that can be generalized to a variety of downstream tasks. However, since these VFMs are generally trained with natural images, they exhibit certain biases in RS applications \cite{ji2024segment}. Considering spectral and temporal characteristics of RSIs, several RS foundation models (FMs) have been developed, including GFM \cite{mendieta2023gfm}, SpectralGPT \cite{hong2024spectralgpt} and SkySense \cite{guo2024skysense}. These FMs enables training-free feature embedding on multi-spectral, multi-temporal, and multi-modal RS data, thereby supporting a variety of downstream tasks including CD. However, since these FMs are typically trained with the context intrinsic to RSIs, they do not consider the specific application context of CD tasks. Consequently, employing these models for CD still necessitates incorporating CD-specific modules and performing fully supervised fine-tuning.

Considering that FMs contain implicit knowledge of the image content, several recent methods have explored employing FMs to achieve sample-efficient CD. In \cite{ding2024samcd}, VFMs are adapted to the RS domain using a semantic latent aligning technique, demonstrating their sample efficiency. Fig.\ref{fig.VFM_CD} presents an overview of this approach, where the latent are aligned via temporal consistency regularization. In \cite{li2024new},

a side-adaption framework is proposed to inject the VFM knowledge into CD models. In \cite{wang2023cs}, SAM is utilized to generate pseudo labels from vague change maps used as prompts. In \cite{zheng2024segment}, zero-shot CD is achieved by measuring the similarity of SAM-encoded features. In \cite{chen2024change}, Chen et al. employed SAM to achieve unsupervised CD between optical images and map data. Dong et al. \cite{dong2024changeclip} utilized CLIP to learn visual-language representations to improve CD accuracy.

\begin{table*}
    \centering
    \caption{Comparison of SOTA accuracy in CD obtained with different sample-efficient methodologies. 'Sup.': supervision type, 'ext.': external data, 'FT': fine-tuning. It should be noted that the experimental settings exhibit variations across different studies in the literature.}\label{Table.SOTA}
    \resizebox{1\linewidth}{!}{%
    \begin{tabular}{l|c|r|cc|ccc}
        \toprule
        \multirow{2}*{Sup.} & \multirow{2}*{Dataset} & \multirow{2}*{Method} & \multirow{2}*{Training data used} & \multirow{2}*{Training label used} & \multicolumn{3}{c}{Accuracy Metrics} \\ \cline{6-8}
        ~ & ~ & ~ & ~ & ~ & OA (\%) & IoU (\%) & $F_1$ (\%) \\ 
        \midrule
        \multirow{12}*{FSCD} & \multirow{4}*{Levir} & BIT \cite{chen2021remote} & 100\% & 100\% & 98.92 & 80.68 & 89.31 \\ 
        ~ & ~ & SAM-CD\cite{ding2024samcd} & 100\% & 100\% & 99.14 & 84.26 & 91.68 \\ 
        ~ & ~ & ScratchF. \cite{noman2024remote} & 100\%  & 100\% & 99.16 & 84.63 & 91.68 \\
        ~ & ~ & Changer\cite{Liky2023Changer} & 100\% & 100\% & --- & --- & 92.06 \\ 
        \cline{2-8}
        ~ & \multirow{3}*{WHU} & BIT \cite{chen2021remote} & 100\% & 100\% & 98.75 & 72.39 & 83.95  \\
        ~ & ~ & ScratchF. \cite{noman2024remote} & 100\%  & 100\% & 99.37 & 84.97 & 91.87 \\
        ~ & ~ & SAM-CD \cite{ding2024samcd} & 100\%  & 100\%  & 99.60 & 91.15 & 95.37  \\
        \cline{2-8}
        ~ & \multirow{5}*{OSCD} & FC-Siam-conc \cite{daudt2018fully} & 100\%  & 100\% & 94.07 & --- & 45.20 \\
        ~ & ~ & FC-Siam-diff \cite{daudt2018fully} & 100\%  & 100\% & 94.86 & --- & 48.86 \\
        ~ & ~ & FC-EF \cite{daudt2018fully} & 100\%  & 100\% & 94.23 & --- & 48.89 \\
        ~ & ~ & ScratchF. \cite{noman2024remote} & 100\%  & 100\% & 97.33 & 40.22 & 57.37 \\
        ~ & ~ & FC-EF-Res \cite{daudt2019multitask} & 100\%  & 100\% & 95.34 & --- & 59.20 \\
        \hline
        \multirow{12}*{SMCD} & \multirow{6}*{Levir} & ECPS\cite{yang2024ecps} & \textcolor{teal}{5\%} \textcolor{cyan}{10\%} \textcolor{orange}{20\%} \textcolor{magenta}{40\%} & \textcolor{teal}{5\%} \textcolor{cyan}{10\%} \textcolor{orange}{20\%} \textcolor{magenta}{40\%} & \textcolor{teal}{98.59} \textcolor{cyan}{98.74} \textcolor{orange}{98.70} \textcolor{magenta}{98.85} & \textcolor{teal}{75.56} \textcolor{cyan}{77.63} \textcolor{orange}{78.06} \textcolor{magenta}{79.30} & \textcolor{teal}{86.06} \textcolor{cyan}{87.40} \textcolor{orange}{87.68} \textcolor{magenta}{88.46} \\         
        ~ & ~ & ST-RCL\cite{zhang2023joint} & \textcolor{teal}{5\%} \textcolor{cyan}{10\%} \textcolor{orange}{20\%} \textcolor{magenta}{40\%} & \textcolor{teal}{5\%} \textcolor{cyan}{10\%} \textcolor{orange}{20\%} \textcolor{magenta}{40\%} & --- & --- & \textcolor{teal}{87.11} \textcolor{cyan}{88.75} \textcolor{orange}{89.46} \textcolor{magenta}{89.77} \\ 
        ~ & ~ & STCRNet\cite{wang2024stcrnet} & \textcolor{teal}{5\%} \textcolor{cyan}{10\%} \textcolor{orange}{20\%} \textcolor{magenta}{40\%} & \textcolor{teal}{5\%} \textcolor{cyan}{10\%} \textcolor{orange}{20\%} \textcolor{magenta}{40\%} & --- & \textcolor{teal}{80.65} \textcolor{cyan}{82.23} \textcolor{orange}{82.98} \textcolor{magenta}{83.48} & \textcolor{teal}{89.29} \textcolor{cyan}{90.25} \textcolor{orange}{90.70} \textcolor{magenta}{91.00} \\ 
        ~ & ~ & UniMatch\cite{yang2023revisiting} & \textcolor{teal}{5\%} \textcolor{cyan}{10\%} \textcolor{orange}{20\%} \textcolor{magenta}{40\%} & \textcolor{teal}{5\%} \textcolor{cyan}{10\%} \textcolor{orange}{20\%} \textcolor{magenta}{40\%} & ~ & \textcolor{teal}{80.88} \textcolor{cyan}{81.73} \textcolor{orange}{82.04} \textcolor{magenta}{82.25} & \textcolor{teal}{89.43} \textcolor{cyan}{89.95} \textcolor{orange}{90.13} \textcolor{magenta}{90.26} \\
        ~ & ~ & C2F-SemiCD\cite{han2024c2fsemicd} & \multicolumn{1}{l}{\textcolor{teal}{5\%} \textcolor{cyan}{10\%} \textcolor{orange}{20\%}}  & \multicolumn{1}{l|}{\textcolor{teal}{5\%} \textcolor{cyan}{10\%} \textcolor{orange}{20\%}} & \multicolumn{1}{l}{\textcolor{teal}{98.99} \textcolor{cyan}{99.08} \textcolor{orange}{99.12}} & \multicolumn{1}{l}{\textcolor{teal}{81.76} \textcolor{cyan}{83.15} \textcolor{orange}{83.75}} & \multicolumn{1}{l}{\textcolor{teal}{89.97} \textcolor{cyan}{90.80} \textcolor{orange}{91.16}} \\ 
        ~ & ~ & ISCDNet\cite{zuo2024robust} & \textcolor{teal}{5\%} \textcolor{cyan}{10\%} \textcolor{orange}{20\%} \textcolor{magenta}{40\%} & \textcolor{teal}{5\%} \textcolor{cyan}{10\%} \textcolor{orange}{20\%} \textcolor{magenta}{40\%} & --- & \textcolor{teal}{81.84} \textcolor{cyan}{82.34} \textcolor{orange}{82.53} \textcolor{magenta}{83.58} & \textcolor{teal}{90.01} \textcolor{cyan}{90.32} \textcolor{orange}{90.43} \textcolor{magenta}{91.06} \\ 
        \cline{2-8}
        ~ & \multirow{5}*{WHU} & UniMatch \cite{yang2023revisiting} & \textcolor{teal}{5\%} \textcolor{cyan}{10\%} \textcolor{orange}{20\%} \textcolor{magenta}{40\%} & \textcolor{teal}{5\%} \textcolor{cyan}{10\%} \textcolor{orange}{20\%} \textcolor{magenta}{40\%} & --- & \textcolor{teal}{75.15} \textcolor{cyan}{77.30} \textcolor{orange}{81.64} \textcolor{magenta}{82.13} & \textcolor{teal}{85.81} \textcolor{cyan}{87.20} \textcolor{orange}{90.95} \textcolor{magenta}{91.26}  \\ 
        ~ & ~ & STCRNet \cite{wang2024stcrnet} & \textcolor{teal}{5\%} \textcolor{cyan}{10\%} \textcolor{orange}{20\%} \textcolor{magenta}{40\%} & \textcolor{teal}{5\%} \textcolor{cyan}{10\%} \textcolor{orange}{20\%} \textcolor{magenta}{40\%} & --- & \textcolor{teal}{77.03} \textcolor{cyan}{81.91} \textcolor{orange}{83.40} \textcolor{magenta}{83.93} & \textcolor{teal}{87.03} \textcolor{cyan}{90.06} \textcolor{orange}{90.95} \textcolor{magenta}{91.26}  \\ 
        ~ & ~ &  ST-RCL \cite{zhang2023joint} & \textcolor{teal}{5\%} \textcolor{cyan}{10\%} \textcolor{orange}{20\%} \textcolor{magenta}{40\%} & \textcolor{teal}{5\%} \textcolor{cyan}{10\%} \textcolor{orange}{20\%} \textcolor{magenta}{40\%} & --- & --- & \textcolor{teal}{87.80} \textcolor{cyan}{88.00} \textcolor{orange}{89.29} \textcolor{magenta}{83.84}  \\ 
        ~ & ~ & C2F-SemiCD \cite{han2024c2fsemicd} & \multicolumn{1}{l}{\textcolor{teal}{5\%} \textcolor{cyan}{10\%} \textcolor{orange}{20\%}}  & \multicolumn{1}{l|}{\textcolor{teal}{5\%} \textcolor{cyan}{10\%} \textcolor{orange}{20\%}} & \multicolumn{1}{l}{\textcolor{teal}{98.87} \textcolor{cyan}{98.94} \textcolor{orange}{99.23}} & \multicolumn{1}{l}{\textcolor{teal}{79.14} \textcolor{cyan}{79.50} \textcolor{orange}{81.93}} & \multicolumn{1}{l}{\textcolor{teal}{88.35} \textcolor{cyan}{88.58} \textcolor{orange}{90.07} } \\ 
        ~ & ~ & ISCDNet \cite{zuo2024robust} & \textcolor{teal}{5\%} \textcolor{cyan}{10\%} \textcolor{orange}{20\%} \textcolor{magenta}{40\%} & \textcolor{teal}{5\%} \textcolor{cyan}{10\%} \textcolor{orange}{20\%} \textcolor{magenta}{40\%} & --- & \textcolor{teal}{81.48} \textcolor{cyan}{82.59} \textcolor{orange}{83.72} \textcolor{magenta}{85.18} & \textcolor{teal}{89.80} \textcolor{cyan}{90.46} \textcolor{orange}{91.14} \textcolor{magenta}{92.00}  \\ 
        \cline{2-8}
        ~ & OSCD & ECPS \cite{yang2024ecps} & \textcolor{teal}{5\%} \textcolor{cyan}{10\%} \textcolor{orange}{20\%} \textcolor{magenta}{40\%} & \textcolor{teal}{5\%} \textcolor{cyan}{10\%} \textcolor{orange}{20\%} \textcolor{magenta}{40\%} & \textcolor{teal}{87.12} \textcolor{cyan}{88.13} \textcolor{orange}{88.59} \textcolor{magenta}{88.98} & \textcolor{teal}{37.05} \textcolor{cyan}{37.69} \textcolor{orange}{40.31} \textcolor{magenta}{41.44} & \textcolor{teal}{54.07} \textcolor{cyan}{54.75} \textcolor{orange}{57.46} \textcolor{magenta}{58.60}  \\ 
        \hline
        \multirow{5}*{WSCD} & \multirow{3}*{Levir} & ICR-MJS\cite{dai2023siamese} & 100\% & image label & --- & 67.41 & 50.84 \\ 
        ~ & ~ & KD-MSI\cite{lu2024weakly} & 100\% & image label & 93.9 & 64.9 & 74.9 \\
        ~ & ~ & CARGNet\cite{fang2023point} & 100\% & point label & 98.28 & 72.13 & 83.81 \\ 
        \cline{2-8}
        ~ & \multirow{2}*{WHU} & ICR-MJS \cite{dai2023siamese} & 100\% & image label & --- & 65.09 & 78.86  \\ 
        ~ & ~ & KD-MSI \cite{lu2024weakly} & 100\% & image label & 99.7 & 76.9 & 85.4  \\ 
        \cline{2-8}
        ~ & OSCD & FCD-GAN \cite{wu2023fully} & 100\% & box label & 91.38 & 21.28 & 35.08  \\ 
        \hline
        \multirow{8}*{SSCD} & \multirow{3}*{Levir} & LGPNet \cite{wang2023self} & 100\% + ext. & 1\% (FT) & --- & 46.13 & 62.09 \\
        ~ & ~ & DST-VGG \cite{zheng2024detail} & 100\% & 100\% (FT) & 99.21 & 85.44 & 92.15 \\ 
        ~ & ~ & RECM \cite{zhang2023selfsupervised} & 100\% + ext.  & 100\% (FT) & --- & --- & 92.77\\         
        \cline{2-8}
	~ & \multirow{2}*{WHU} & GLCL \cite{jiang2023self} & 100\% & 100\% (FT) & --- & 90.29 & 90.54 \\
        ~ & ~ & DST-VGG \cite{zheng2024detail} & 100\% & 100\% (FT) & 99.64 & 90.34 & 95.69 \\
        \cline{2-8}
        ~ & \multirow{3}*{OSCD} & PixSSLs \cite{chen2022self} & 100\% + ext. & 0\% & 95.70 & --- & 53  \\ 
        ~ & ~ & DK-SSCD \cite{yan2023domain} & 100\% & 0\% & 95.54 & --- & 55.69  \\ 
        ~ & ~ & TD-SSCD \cite{qu2023tdsscd} & 100\% & 100\% (FT) & 95.38 & --- & 72.11  \\ 
        \hline
        \multirow{9}*{UCD} & \multirow{4}*{Levir} & 
        Anychange\cite{zheng2024segment} & 0\% & 0\% & --- & --- & 23.0 \\
        ~ & ~ & DSFA \cite{Du2019Unsupervised} & 0\% & 0\% & 77.33 & --- & 47.65 \\
        ~ & ~ & DCVA \cite{saha2019unsupervised} & 0\% & 0\% & 84.75 & --- & 52.89 \\
          ~ & ~ & SCM \cite{noh2022unsupervised} & 100\% & 0\% & 88.80 & --- & 62.80 \\
        \cline{2-8}
        ~ & \multirow{5}*{OSCD} & DCVA \cite{saha2019unsupervised} & 0\% & 0\% & 91.6 & --- & 24.5 \\
        ~ & ~ & KPCA-MNet \cite{wu2021unsupervised} & 0\% & 0\% & --- & --- & 30.2 \\
        ~ & ~ & FLCG \cite{Mall2022Change} & 100\% & 0\% & --- & --- & 32.1 \\
        ~ & ~ & DMLCD \cite{bandara2023deep} & 100\% & 0\% & 95.8 & --- & 32.5 \\
        ~ & ~ & DSFA \cite{Du2019Unsupervised} & 0\% & 0\% & 92.63 & --- & 35.85 \\

        \bottomrule
    \end{tabular}}
\end{table*}

\begin{table*}[!ht]
    \centering
    \caption{Statistical overview of the benchmark CD datasets presented in Table \ref{Table.SOTA}.}\label{Table.Datasets}
    \resizebox{1\linewidth}{!}{%
    \begin{tabular}{c|c|r|r|r|r|ccccc}
    \toprule
        \multirow{2}*{Datasets} & \multirow{2}*{Resolution} & \multirow{2}*{Image size} & Image & Change & Change & \multicolumn{5}{c}{Highest $F_1$ (\%)} \\
        \cline{7-11}
        & & & Pairs & Pixels & Instances &  FSCD & SMCD (5\%)  & WSCD & SSCD (w/o. FT) & UCD \\
        \hline
        Levir & 0.5m & 1024×1024 & 637 & 30,913,975 & 31,333 & 92.06 \cite{Liky2023Changer} & 90.01\cite{zuo2024robust} & 74.9 \cite{dai2023siamese} &   & 62.80 \cite{noh2022unsupervised} \\ \hline
        WHU & 0.3m & 32,507×15,354 & 1 & 21,352,815 & 2297 & 95.37 \cite{ding2024samcd} & 89.80 \cite{zuo2024robust} & 85.4 \cite{lu2024weakly} & --- & --- \\ \hline
        OSCD & 10m & 600×600 & 24 & 148,069 & 1048 & 59.20  \cite{daudt2019multitask} & 54.07 \cite{yang2024ecps} & 35.08 \cite{wu2023fully} & 55.69 \cite{yan2023domain} & 35.85 \cite{Du2019Unsupervised} \\
    \bottomrule
    \end{tabular}}
\end{table*}

\subsection{Comparison of Accuracy}  \label{sc3-e}

To elucidate the efficacy of the sample-efficient DL methodologies discussed, Table \ref{Table.SOTA} presents a comparative analysis of the SOTA accuracy obtained on several benchmark CD datasets. The accuracy metrics include overall accuracy (OA), intersection over union (IoU), and $F_1$, which are common in BCD. To facilitate comparison between different types of supervision, we select the most frequently used datasets in various tasks, including Levir \cite{Chen2020}, WHU \cite{ji2018fully}, and OSCD \cite{daudt2018urban}. It is important to acknowledge that there are significant variations in the experimental configurations of the methods being compared, a concern raised in \cite{corley2024change}. Therefore, this table is intended solely to provide an intuitive assessment of the accuracy of SOTA.

To facilitate a comprehensive understanding of the training samples utilized across various methods, Table \ref{Table.Datasets} presents the metadata of each CD benchmark. Overall, Levir and WHU are two VHR datasets with large image size and rich change samples. In contrast, OSCD has lower resolution and contains less training samples. To present an intuitive comparison of the SOTA accuracy across various learning paradigms, Table \ref{Table.Datasets} also summarizes the highest $F_1$ scores achieved in each dataset.

Tables \ref{Table.SOTA} and \ref{Table.Datasets} clearly demonstrate that the accuracy of CD is highly dependent on the level of supervision during the training process. First, the CD accuracy on the Levir and WHU datasets is significantly higher than that on the OSCD dataset. This disparity is attributed to the richer set of change samples and the finer spatial resolution present in the Levir and WHU datasets. Second, the accuracy of FSCD and SSCD with fine-tuning (FT) is higher than that of SMCD, WSCD, UCD and SSCD without FT. Notably, the SSCD with FT marginally surpasses FSCD, which can be attributed to its extensive pre-training that effectively utilizes the image contexts as extra supervisions. This observed accuracy hierarchy aligns with the strong-to-weak supervision level in the different learning paradigms, as illustrated in Fig.\ref{fig.learning_types}.

SMCD achieves the highest accuracy among sample-efficient CD approaches. Recent advances in SMCD ensure remarkably high accuracy with only a small proportion of training samples. For example, utilizing only 5\% of the training data, the SOTA SMCD methods only see a minor $F_1$ reduction of 2\% on the Levir and 0.6\% on the OSCD datasets. However, it is important to note that even with this small portion of training data, SMCD still requires a substantial number of change samples. Based on the number of change instances detailed in Table \ref{Table.SOTA} and through a rough estimate, SMCD typically requires more than 100 change samples on the Levir and WHU datasets.

The accuracy of WSCD is significantly influenced by the level of supervision applied. Compared to image-level labels, employing spatial labels (such as box or point labels) for training WSCD algorithms generally results in superior accuracy. For example, as tested on the Levir dataset, point label-supervised CD approach \cite{fang2023point} has an advantage of exceeding 30\% in $F_1$ compared to approaches that utilize image labels. Regarding image label-supervised SMCD, while a relatively high accuracy is attained (particularly on VHR datasets), it is important to note that training is carried out using patch labels rather than a complete RSI. As reported in \cite{lu2024weakly} and \cite{dai2023siamese}, image labels are assigned to each pair of patches with 256$\times$256 pixels. Therefore, this type of SMCD still necessitates a certain degree of human intervention.

SSCD can be employed as either an approach to achieve label-free learning of change representations, or merely as a pretraining technique to initialize the DNN parameters. SSCD without FT is challenging, since the image contexts utilized in self-supervised learning are independent of the application contexts inherent in CD tasks. Most literature works adopt the latter strategy, that is, pre-training through self-supervision and fine-tuning with all available change samples. This strategy yields substantial accuracy improvements over the vanilla FSCD. The improvements are particularly significant on the OSCD dataset (up to 12\% in $F_1$ \cite{qu2023tdsscd}), which can be attributed to the scarcity of training samples within this dataset. 

Meanwhile, SSCD without FT can be regarded as a distinct subset of UCD that utilizes self-supervised learning techniques. Most literature studies on UCD and label-free SSCD have been conducted on medium resolution datasets such as OSCD. They are commonly adopted for analysis of satellite images such as those collected by Sentinel and Landsat. Due to the fact that numerous experiments are performed on non-open benchmarks, it is challenging to assess the level of accuracy, and hence, these results are not presented in Table \ref{Table.SOTA}. The highest metrics obtained on the OSCD dataset are 92.63\% in OA and 35.85\% in $F_1$ \cite{Du2019Unsupervised}, exhibiting a reduction exceeding 23\% in $F_1$ relative to FSCD. UCD (or label-free SSCD) is more challenging when applied to VHR datasets due to the increased spatial complexity. A reduction of approximately 30\% in $F_1$ is noted when applied to the Levir dataset. One of the zero-shot CD approaches, Anychange \cite{zheng2024segment}, obtains an accuracy of 24.5\% in $F_1$, highlighting a substantial gap for further advancements.

In summary, sample-efficient CD methods have greatly reduced the dependence on a large volume of training samples, thereby achieving relatively high accuracy with a reduced number of samples or the utilization of weak labels. However, training CD algorithms without labels or using a very low level of supervision remains a challenge.

\section{challenges and Future Trends}\label{sec_future}
Despite the advanced methodologies and techniques developed, training DNNs for CD with very few samples remains challenging. This section presents an analysis of the remaining challenges and bottleneck problems in applying sample-efficient CD algorithms, along with a prospective overview of future developments in the field.

\subsection{Challenges}

Sample-efficient CD still encounters considerable challenges in mitigating the data dependency and in generalizing insights across diverse datasets without necessitating extensive fine-tuning. In the followings we analyse several principal obstacles.

\subsubsection{Domain adaptability}

RSIs collected by different sensors and platforms exhibit considerable variability in spatial resolution, imaging scale, and spectral patterns. Conventional machine learning methods derive different levels of analysis on pixel spectrals, local textures, and object contexts \cite{wen2021change, lv2022land}. Despite the capability of DL to facilitate end-to-end modeling of multilevel change patterns, these approaches still suffer from severe accuracy degradation when dealing with data from other domains. Although there are heterogeneous CD methods, they are constrained by trained domain transitions and face challenges in obtaining domain-invariant change representations.

The major reasons are two-fold:
i) domain-specific network architectures. DL-based CD methods employ diversified DL techniques to perform intricate analysis on the informative attributes in different RS data. For instance, some methods employ spectral attention \cite{hu2022hypernet} and superpixel GNNs \cite{qu2023feature} for hyperspectral CD, while some other methods introduce low-level supervision \cite{peng2019end, lu2020weakly} and geometric perturbations \cite{wang2024stcrnet} for CD in VHR RSIs. Although these designs yield significant accuracy enhancements within the training domains, they present substantial challenges when it comes to generalizing to novel domains.
ii) Domain-coupled change learning. Typical CD approaches learn mappings of difference patterns specific to training domains. Although there exist domain-invariant change representation methods \cite{sublime2019automatic}, they also neglect the intrinsic semantic transition mechanism in CD. Consequently, the resultant models struggle to differentiate between specific semantic changes and unknown domain variations.

\subsubsection{Spatial and Temporal Complexity}

Recent advancements in Earth Observation technologies enable dense time-series monitoring through the deployment of surveillance satellites and small satellite constellations. The improvement in temporal resolution benefits applications that require frequent observations, including environmental monitoring, urban management, and disaster alarm. However, DL-based analysis on time series CD is still in an early exploration phase, especially for the analysis of long-time series of HR images. Conventional methods address TSCD as a multidate LCLU classification task or analyze the trajectory of multi-temporal images \cite{stahl2023automated}. This neglects the spatio-temporal context in HR data and may result in false alarms due to temporal variations (such as temporary occlusions and seasonal changes). 

Additionally, few studies address the spatial misalignment that often occurs in multi-source RSIs. Observation platforms such as UAVs and surveillance satellites offer quick access to regions of interest due to shorter revisiting. However, these platforms differ greatly in imaging angles and geometric distortions \cite{shen2021s2looking}. Most CD studies require costly and time-consuming preprocessing operations to ensure strict spatial consistency, thereby constraining their applicability. To expand the applicability of CD techniques, there is a research gap in developing sample-efficient methodologies to address the spatial and temporal complexity in CD.

\subsubsection{Unseen changes} Sample-efficient CD requires identification of changes that are absent from the training data. Unseen changes can be classified into two distinct types: (i) change instances that exhibit novel appearances yet remain within the established categories, a frequent occurrence in SMCD due to constrained usage of training data; and (ii) novel categories of changes that remain undefined, a common situation in SSCD and UCD while transferring domain knowledge into new datasets.

In BCD, the major challenge lies in distinguishing semantic changes amid temporal variations; whereas in MCD/SCD, the difficulties additionally encompass the identification of novel change categories. These challenges can be further amplified while encountering the previously mentioned obstacles including domain gap and spatio-temporal complexity. Generalization of CD insights into wider RS applications requires a profound understanding of the semantic transitions, as well as comprehension of the specific application contexts.

\subsection{Future trends}

\subsubsection{Multi-temporal foundation models}

Recent breakthroughs in generative image synthesis, self-supervision techniques, and VFMs are setting the stage for the next generation of CD algorithms \cite{ding2024samcd, zheng2024segment}. Although variable VFMs \cite{Kirillov2023Segment} and spectral foundation models \cite{hong2024spectralgpt, guo2024skysense} have been established within the domains of computer vision and RS, the development of multi-temporal foundation models (TFMs) is crucial to achieve sample-efficient, sensor-agnostic, and eventually training-free CD within a unimodal framework.

TFMs are designed to capture temporal patterns and dynamic changes across multiple observations and subsequently utilizing the learned temporal knowledge to identify evolving trends. These models are designed to manage complex spatio-temporal dependencies within time-series RSIs, address data heterogeneity, and adapt to varying temporal intervals. They ensure scalability for extensive volumes of RS big data and create universal change representations by seamlessly integrating diverse sensor data across a range of resolutions and scales.

\subsubsection{Few-shot and Zero-shot CD}

As detailed in Sec.\ref{sc3-e}, most literature studies on SMCD still require a considerable number of training samples to achieve accurate results. In real-world applications, collecting change samples is costly, especially when data is scarce or quick responses are necessary. Thus, developing few-shot and zero-shot CD algorithms is critical for deploying CD systems with minimal change samples.

Few-shot learning (FSL) aims to acquire generalized knowledge applicable across various tasks using only a few examples. Most of the literature methods on FSL follow the meta-learning framework proposed in \cite{vinyals2016matching}. This framework mimics the few-shot applicational scenarios, where the network learns to identify novel classes in the unlabeled data (query set) by utilizing the knowledge obtained from a few number of examples in the labeled data (support set). FSL allows DNNs to generalize to novel classes from a minimum of just one example, and has been investigated in the task of semantic segmentation \cite{lang2022learning}. 

Zero-shot learning (ZSL) uses data from known classes to train DNNs, enabling inference on unseen classes. Typical FSL methods map visual and semantic features to a common space for data-independent semantic alignment. Several problem settings have been further derived from ZSL to address various distinct application contexts \cite{rahman2022polarity}. These include: i) transductive ZSL, which uses unlabeled unseen data in training; ii) generalized ZSL, which involves classifying both seen and unseen classes; iii) domain adaptation, which adapts unseen targets to seen source domains; and iv) class-attribute association, which links unsupervised semantics to human-recognizable attributes. 

Integrating FSL and ZSL into CD methods could remove the need for fine-tuning algorithms on target domains. However, the context of CD presents more severe challenges, such as data heterogeneity and reduced density of semantic contexts. Few-shot and zero-shot CD remain to be rarerly explored and requires further research investigations.

\subsubsection{Interactive CD}

In many practical cases, CD is closely associated with the specific application context, such as urban building changes or agricultural monitoring. Conventional DL-based CD implicitly learn these applicational contexts through training samples, which is challenging with scarce change samples. An alternative is to incorporate explicit human interactions to guide the active exploitation of the relevant change information. Two key interaction types are spatial and language interactions.

a. Spatial interactions. In various VFMs, user-generated input, such as points, scribbles, and rectangles, is encoded as spatial prompts to indicate the interesting objects to be extracted/segmented \cite{Kirillov2023Segment}. This approach can be expanded to CD tasks by incorporating bi-temporal annotations to specify the change objects of interest. This depends on the application of WSCD methodologies, which entails parsing weak spatial annotations into dense change predictions. Moreover, to minimize human effort and achieve the capability of 'clicking few and detecting many', the incorporation of SMCD and continual learning techniques \cite{douillard2021plop} is essential. The former facilitate the efficient use of sparse and scarce change annotations, while the latter allows for interactive refining and updating of annotations to specify the desired changes.

b. Language interactions. Recent developments in the fusion of language models with RS data analysis represent a new frontier in CD, offering innovative ways to interact with and interpret CD results. This approach includes: i) change captioning: describing the major changes in multi-temporal RSIs, ii) prompt-driven CD: selectively segment the changes of interested LCLU categories given user prompt such as keywords \cite{dong2024changeclip}, and iii) visual question answering for CD: given questions concerning changes on RSIs, providing detailed and informative language answers. Language-driven CD offers a more intuitive interface between users and CD systems.

\section{Conclusions}

Leveraging limited data to train DNNs with dense parameters has consistently been a bottleneck challenge in the deployment of DL algorithms. Recently, with the ongoing progress in DL methodologies such as image generation, self-supervised learning, and VFMs, there has been a growing increase in research attention towards sample-efficient CD.

CD has consistently been an important visual recognition task in RS applications. It can be classified into BCD, MCD/SCD, and TSCD, based on the granularity of the results and the number of observation dates. Sample-efficient change detection can be categorized into distinct learning paradigms based on the diversity in label forms and quantities. These paradigms encompass four principal types, including SMCD, WSCD, SSCD, and UCD. Each learning setting further derives diverse strategies and technologies specifically designed to overcome the unique challenges presented, which have been systematically reviewed and summarized in Table \ref{Table.Strategy}. Moreover, to facilitate an intuitive comprehension of the SOTA performance in sample-efficient CD, a comparative analysis is performed across various learning settings with regard to change samples and the achieved accuracy. Finally, a critical analysis of the challenges encountered is provided, along with recommendations for potential future research directions.

In conclusion, the exploration of sample-efficient CD is still in an early stage of exploration. Although notable progress has been made in decreasing the dependence on extensive training samples, the challenge of performing CD with very scarce samples persists. There exists a substantial research gap in the development of CD methodologies to tackle more challenging CD scenarios, such as few-shot CD, image label-supervised WSCD, unsupervised CD (UCD), non-fine-tuned SSCD, and ultimately zero-shot CD.

\section*{Acknowledgement}
This work was supported by the National Natural Science Foundation of China under Grant 42201443, Grant 42271350, and also supported by the International Partnership Program of the Chinese Academy of Sciences under Grant No.313GJHZ2023066FN. Danfeng Hong is the corresponding author.

\bibliographystyle{IEEEtran}
\bibliography{refs}

\vskip -2\baselineskip plus -1fil

\begin{IEEEbiographynophoto}
    {Lei Ding} (dinglei14@outlook.com) received his MS’s degree in Photogrammetry and Remote Sensing from the Information Engineering University (Zhengzhou, China), and his PhD (cum laude) in Communication and Information Technologies from the University of Trento (Trento, Italy). He is currently a Lecturer at the Information Engineering University. Since 2024, he has been a Post-doctoral Fellow at the Aerospace Information Research Institute, Chinese Academy of Sciences, Beijing, China. His research interests are related to the intelligent interpretation of remote sensing data.
\end{IEEEbiographynophoto}

\vskip -2\baselineskip plus -1fil

\begin{IEEEbiographynophoto}
    {Danfeng Hong} (IEEE Senior Member, hongdf@aircas.ac.cn) received the Dr. -Ing degree (summa cum laude) from the Signal Processing in Earth Observation (SiPEO), Technical University of Munich (TUM), Munich, Germany, in 2019. 

    Since 2022, he has been a Full Professor with the Aerospace Information Research Institute, Chinese Academy of Sciences. His research interests include Artificial Intelligence, Multimodal, Foundation Models, and Earth Observation. 
    
    Dr. Hong serves as an Associate Editor for the IEEE Transactions on Image Processing (TIP) and the IEEE Transactions on Geoscience and Remote Sensing (TGRS). He is also an Editorial Board Member for Information Fusion and the ISPRS Journal of Photogrammetry and Remote Sensing. He has received several prestigious awards, including the Jose Bioucas Dias Award (2021) and Paul Gader Award (2024) at WHISPERS for outstanding papers, respectively, the Remote Sensing Young Investigator Award (2022), the IEEE GRSS Early Career Award (2022), and the ``2023 China’s Intelligent Computing Innovators'' award (the only recipient in AI for Earth Science) by MIT Technology Review (2024). He has been recognized as a Highly Cited Researcher by Clarivate Analytics in 2022, 2023, and 2024.
\end{IEEEbiographynophoto}

\vskip -2\baselineskip plus -1fil

\begin{IEEEbiographynophoto}
    {Maofan Zhao} (mfzhao1998@163.com) is currently pursuing a Ph.D. degree at the Aerospace Information Research Institute, Chinese Academy of Sciences, Beijing, China. From 2021 to 2023, he was a Visiting Ph.D. student at the University of Trento, Trento, Italy. His research interests include remote sensing image analysis, deep learning and urban remote sensing.
\end{IEEEbiographynophoto}

\vskip -2\baselineskip plus -1fil

\begin{IEEEbiographynophoto}
    {Hongruixuan Chen} (IEEE Student Member, qschrx@gmail.com) received the M.E. degree in photogrammetry and remote sensing from State Key Laboratory of Information Engineering in Surveying, Mapping, and Remote Sensing, Wuhan University, Wuhan, China, in 2022. He is currently pursuing the Ph.D. degree with the Graduate School of Frontier Science, The University of Tokyo, Chiba, Japan. His research fields include deep learning, domain adaptation, and multimodal remote sensing image interpretation and analysis.
\end{IEEEbiographynophoto}

\vskip -2\baselineskip plus -1fil

\begin{IEEEbiographynophoto}
    {Chenyu Li} (lichenyu@seu.edu.cn) received the B.S. and M.S. degrees from the School of Transportation, Southeast University, Nanjing, China, in 2018 and 2021, respectively. 

    She is currently pursuing her Ph.D. degree from the Department of Mathematics at Southeast University, Nanjing, China, and is also a joint Ph.D. student at the Aerospace Information Research Institute, Chinese Academy of Sciences, Beijing, China. Her research interests include interpretable artificial intelligence, big Earth data forecasting, foundation models, and hyperspectral imaging.
\end{IEEEbiographynophoto}

\vskip -2\baselineskip plus -1fil

\begin{IEEEbiographynophoto}
    {Jie Deng} received the Ph.D. degree in agronomy from China Agricultural University, Beijing, China, in 2023. Since 2023, he has been working as a Post-Doctoral Researcher with the College of Plant Protection, China Agricultural University. His main research areas include plant disease epidemiology, quantitative inversion of crop diseases based on imaging remote sensing, and yield prediction.
\end{IEEEbiographynophoto}

\vskip -2\baselineskip plus -1fil

\begin{IEEEbiographynophoto}
    {Naoto Yokoya} (IEEE Member, naoto.yokoya@riken.jp) is currently an Associate Professor at The University of Tokyo, Japan, and he leads the Geoinformatics Team at the RIKEN Center for Advanced Intelligence Project, Tokyo. His research is focused on the development of image processing, data fusion, and machine learning algorithms for understanding remote sensing images, with applications to disaster management and environmental assessment.
\end{IEEEbiographynophoto}

\vskip -2\baselineskip plus -1fil

\begin{IEEEbiographynophoto}
    {Lorenzo Bruzzone} (IEEE Fellow, lorenzo.bruzzone@unitn.it) received his M.S. degree in electronic engineering (summa cum laude) and his Ph.D. degree in telecommunications from the University of Genoa, Italy, in 1993 and 1998, respectively. Currently, he is with the Department of Information Engineering and Computer Science, University of Trento, Italy. Dr. Bruzzone is the founder and the director of the Remote Sensing Laboratory in the Department of Information Engineering and Computer Science, University of Trento. His research interests include remote sensing, radar and SAR, signal processing, machine learning, and pattern recognition.
\end{IEEEbiographynophoto}

\vskip -2\baselineskip plus -1fil

\begin{IEEEbiographynophoto}
    {Jocelyn Chanussot} (IEEE Fellow, jocelyn.chanussot@inria.fr) received the M.Sc. degree in electrical engineering from the Grenoble Institute of Technology (Grenoble INP), Grenoble, France, in 1995, and the Ph.D. degree from the Université de Savoie, Annecy, France, in 1998. From 1999 to 2023, he has been with Grenoble INP, where he was a Professor of signal and image processing. He is currently a Research Director with INRIA, Grenoble. His research interests include image analysis, hyperspectral remote sensing, data fusion, machine learning, and artificial intelligence. He has been a visiting scholar at Stanford University (USA), KTH (Sweden) and NUS (Singapore). Since 2013, he is an Adjunct Professor of the University of Iceland. In 2015-2017, he was a visiting professor at the University of California, Los Angeles (UCLA).  He holds the AXA chair in remote sensing and is an Adjunct professor at the Chinese Academy of Sciences, Aerospace Information research Institute, Beijing, China.

    Dr. Chanussot is the founding President of the IEEE Geoscience and Remote Sensing French chapter (2007-2010) which received the 2010 IEEE GRSS Chapter Excellence Award. He was the Vice-President of the IEEE Geoscience and Remote Sensing Society, in charge of meetings and symposia (2017-2019). He is an Associate Editor for the IEEE Transactions on Geoscience and Remote Sensing, the IEEE Transactions on Image Processing, and the Proceedings of the IEEE. He was the Editor-in-Chief of the IEEE Journal of Selected Topics in Applied Earth Observations and Remote Sensing (2011-2015). In 2014 he served as a Guest Editor for the IEEE Signal Processing Magazine. He is a Fellow of the IEEE, an ELLIS Fellow, a Fellow of AAIA, a member of the Institut Universitaire de France (2012-2017,) and a Highly Cited Researcher (Clarivate Analytics/Thomson Reuters, since 2018).
\end{IEEEbiographynophoto}

\end{document}